%% file: main.tex
\documentclass{article}
\usepackage[final,nonatbib]{neurips_2020}
\pdfoutput=1

\usepackage{booktabs} 
\usepackage{times}
\usepackage{graphicx}
\usepackage{sidecap}
\usepackage{wrapfig}

\title{Beyond the Mean-Field: Structured Deep Gaussian Processes Improve the Predictive Uncertainties}

\newcommand\new[1]{\color{black}#1}
\newcommand*{\affaddr}[1]{#1} 
\newcommand*{\affmark}[1][*]{\textsuperscript{#1}}

\author{%
	Jakob Lindinger\affmark[1,2,3], David Reeb\affmark[1], Christoph Lippert\affmark[2,3], Barbara Rakitsch\affmark[1]\\
	\affaddr{\affmark[1]Bosch Center for Artificial Intelligence, Renningen, Germany}\\
	\affaddr{\affmark[2]Hasso Plattner Institute, Potsdam, Germany}\\
	\affaddr{\affmark[3]University of Potsdam, Germany}\\
	\{jakob.lindinger, david.reeb, barbara.rakitsch\}@de.bosch.com, christoph.lippert@hpi.de
}

\usepackage{multirow}
\usepackage{mathtools}
\usepackage{amsmath}
\usepackage{amssymb}
\usepackage{amsthm}
\usepackage{algorithm}
\usepackage{algorithmicx}
\usepackage[noend]{algpseudocode}
\usepackage{xcolor} 
\usepackage{enumerate}
\newcommand{\gauss}[3]{\mathcal{N}\left( #1 \middle| #2, #3 \right)}

\newtheorem{theorem}{Theorem}
\newtheorem{lemma}[theorem]{Lemma}

\newcommand{\SM}[2]{S_M^{#1,#2}}
\newcommand{\Sn}[2]{\widetilde{S}_n^{#1,#2}}

\newcommand{\lemdiagK}[2]{\emph{diag}(\widetilde{\mathcal{K}}_{#1}^{1:#2})}
\newcommand{\Sninv}{\left( \widetilde{S}_n^{1:l-1,1:l-1} \right)^{-1}}
\newcommand{\SigM}[2]{{}^l\hat{\Sigma}_M^{#1,#2}}
\newcommand{\SignM}[2]{{}^l\hat{\Sigma}_{nM}^{#1,#2}}
\newcommand{\SigMn}[2]{\left( \SignM{#1}{#2} \right)^\top}
\newcommand{\muM}[1]{{}^l\hat{\mu}_M^{#1}}
\newcommand{\Signinv}{\left(\hat{\Sigma}_n^{l}\right)^{-1}}
\usepackage{hyperref}

\begin{document}

\maketitle

\input{subfiles/abstract.tex}

\input{subfiles/intro.tex}

\input{subfiles/background.tex}

\input{subfiles/dgp.tex}

\input{subfiles/related_work.tex}

\input{subfiles/experiments.tex}

\input{subfiles/summary.tex}
\input{subfiles/impact.tex}

\input{subfiles/appendix_theory.tex}
\input{subfiles/appendix_theory_intuition.tex}
\input{subfiles/appendix_theory_part2.tex}
\input{subfiles/appendix_theory_part3.tex}
\input{subfiles/appendix_pseudocode.tex}
\input{subfiles/appendix_exp.tex}

\end{document}

%% file: subfiles/abstract.tex
\begin{abstract}
Deep Gaussian Processes learn probabilistic data representations for supervised learning by cascading multiple Gaussian Processes. While this model family promises flexible predictive distributions, exact inference is not tractable. Approximate inference techniques trade off the ability to closely resemble the posterior distribution against speed of convergence and computational efficiency. We propose a novel Gaussian variational family that allows for retaining covariances between latent processes while achieving fast convergence by marginalising out all global latent variables. After providing a proof of how this marginalisation can be done for general covariances, we restrict them to the ones we empirically found to be most important in order to also achieve computational efficiency. We provide an efficient implementation of our new approach and apply it to several benchmark datasets. It yields excellent results and strikes a better balance between accuracy and calibrated uncertainty estimates than its state-of-the-art alternatives.
\end{abstract}

%% file: subfiles/intro.tex
\section{Introduction}
Gaussian Processes (GPs) provide a non-parametric framework for learning distributions over unknown functions from data~\cite{rasmussen2004gaussian}:
As the posterior distribution can be computed in closed-form, they return well-calibrated uncertainty estimates, making them particularly useful in safety critical applications~\cite{amodei2016concrete, reeb2018learning}, Bayesian optimisation~\cite{hebbal2019bayesian, snoek2012practical}, active learning~\cite{zimmer2018safe} or under covariate shift~\cite{snoek2019can}.
However, the analytical tractability of GPs comes at the price of reduced flexibility: 
Standard kernel functions make strong assumptions such as stationarity or smoothness.
To make GPs more flexible, a practitioner would have to come up with hand-crafted features or kernel functions. 
Both alternatives require expert knowledge and are prone to overfitting.

Deep Gaussian Processes (DGPs) offer a compelling alternative since they learn non-linear feature representations in a fully probabilistic manner via GP cascades~\cite{damianou2013dgp}.
The gained flexibility has the drawback that inference can no longer be carried out in closed-form, but must be performed via Monte Carlo sampling~\cite{havasi2018inference}, or approximate inference techniques~\cite{bui2016deep, damianou2013dgp, salimbeni-deisenroth-doubly-stochastic-vi-deep-gp}. 
The most popular approximation, variational inference, 
searches for the best approximate posterior within a pre-defined class of distributions: the variational family~\cite{blei2017variational}.
For GPs, variational approximations often build on the inducing point framework where a small set of global latent variables acts as pseudo datapoints 
summarising the training data~\cite{snelson-spgp, titsias2009variational}.
For DGPs, each latent GP is governed by its own set of inducing variables, which, in general, need not be independent from those of other latent GPs.
Here, we offer a new class of variational families for DGPs taking the following two requirements into account: 
(i) all global latent variables, i.e., inducing outputs, 
can be marginalised out,
(ii) correlations between latent GP models can be captured.
Satisfying (i) reduces the variance in the estimators and is needed for fast convergence~\cite{kingma2015variational} while 
(ii) leads to better calibrated uncertainty estimates~\cite{turner2011two}.

\begin{wrapfigure}{l}{0.415 \textwidth}
\centering
		\includegraphics[width=0.415 \textwidth]{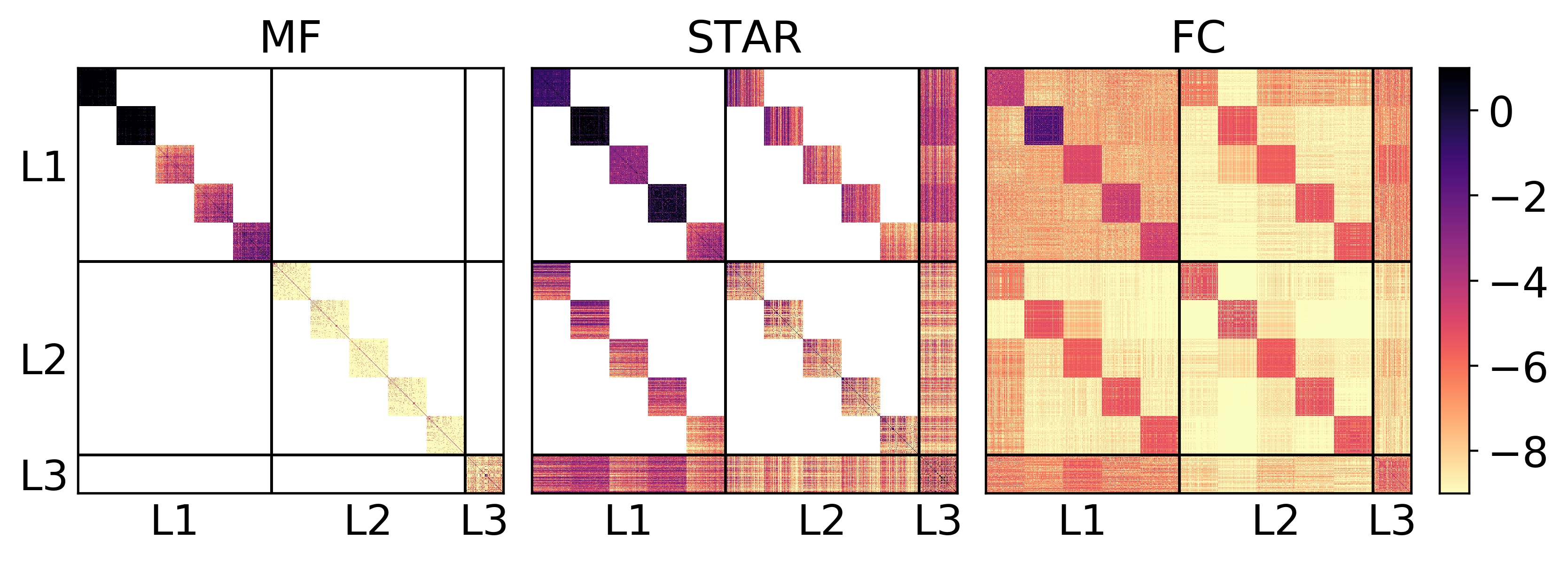}	
	\caption{
	\textbf{Covariance matrices for variational posteriors.
	}
	We used a DGP with 2 hidden layers (L1, L2) of 5 latent GPs each and a single GP in the output layer (L3). The complexity of the variational approximation is increased by allowing for additional dependencies within and across layers in a Gaussian variational family (left: mean-field \cite{salimbeni-deisenroth-doubly-stochastic-vi-deep-gp}, middle: stripes-and-arrow, right: fully-coupled).
	Plotted are natural logarithms of the absolute values of the variational covariance matrices over the inducing outputs.}
	\label{fig:covariance}
\end{wrapfigure}

By using a fully-parameterised Gaussian variational posterior over the global latent variables, we automatically fulfil (ii), and we show in Sec.~\ref{sec:fc}, via a proof by induction, that (i) can still be achieved.
The proof is constructive, resulting in a novel inference scheme for variational families that allow for correlations within and across layers.
The proposed scheme is general and can be used for arbitrarily structured covariances allowing the user to easily adapt it to application-specific covariances, depending on the desired DGP model architecture and on the system requirements with respect to speed, memory and accuracy.
One particular case, in which the variational family is chain-structured, has also been considered in a recent work~\cite{ustyuzhaninov2019compositional}, in which the compositional uncertainty in deep GP models is studied. 

In Fig.~\ref{fig:covariance} (right) we depict exemplary inferred covariances between the latent GPs for a standard deep GP architecture.
In addition to the diagonal blocks, the covariance matrix has visible diagonal stripes in the off-diagonal blocks and an arrow structure.
These diagonal stripes point towards strong dependencies between successive latent GPs, while the arrow structure reflects dependencies between all hidden layers and the output layer.
In Sec.~\ref{sec:afc},  we further propose a scalable approximation to this variational family, which only takes these stronger correlations into account (Fig.~\ref{fig:covariance}, middle).
We provide efficient implementations for both variational families, where we particularly exploit the sparsity and structure of the covariance matrix of the variational posterior.
In Sec.~\ref{sec:experiments}, we show experimentally that the new algorithm works well in practice.
Our approach obtains a better balance between accurate predictions and calibrated uncertainty estimates than its competitors, as we showcase by varying the distance of the test from the training points.

%% file: subfiles/background.tex
\section{Background}
In the following, we introduce the notation and  provide the necessary background on DGP models.
GPs are their building blocks and the starting point of our review.

\subsection{Primer on Gaussian Processes}
In regression problems, the task is to learn a function $ f: {\mathbb{R}}^D\to  {\mathbb{R}}$ that maps a set of $ N $ input points $ x_N = \{x_n\}_{n=1}^N $ to a corresponding set of noisy outputs $ y_N = \{y_n\}_{n=1}^N $.
Throughout this work, we assume iid noise, $ p(y_N|f_N) =  \prod_{n=1}^{N} p(y_n|f_n)$, where $ f_n = f(x_n) $ and $ f_N = \{f_n\}_{n=1}^N $ are the function values at the input points.
We place a zero mean GP prior on the function $ f $, $ f \sim \mathcal{GP}(0,k) $, where $ k:  {\mathbb{R}}^D \times  {\mathbb{R}}^D \to {\mathbb{R}} $ is the kernel function.
This assumption leads to a multivariate Gaussian prior over the function values, $ p(f_N) = \gauss{f_N}{0}{K_{NN}} $ with covariance matrix $ K_{NN} = \{k(x_n,x_{n'})\}_{n,n'=1}^N $.

In preparation for the next section, we introduce a set of  $ M \ll N $ so-called inducing points $ x_M = \{x_m\}_{m=1}^M $ from the input space\footnote{Note that in our notation variables with an index $ m,M\ (n,N) $ denote quantities related to inducing (training) points. This implies for example that $ x_{m=1} $ and $ x_{n=1} $ are in general not the same.}~\cite{snelson-spgp,titsias2009variational}.
From the definition of a GP, the corresponding inducing outputs $ f_M = \{f_m\}_{m=1}^M$, where $f_m = f(x_m) $, share a joint multivariate Gaussian distribution with $ f_N $. We can therefore write the joint density as
$
p(f_N,f_M) = p(f_N|f_M)p(f_M),
$
where we factorised the joint prior into $ p(f_M) = \gauss{f_M}{0}{K_{MM}} $, the prior over the inducing outputs, and the conditional $ p(f_N|f_M) = \gauss{f_N}{\widetilde{K}_{NM}f_M}{\widetilde{K}_{NN}} $ with
\begin{align}
\widetilde{K}_{NM} = K_{NM}\left(K_{MM}\right)^{-1},  
 \quad 
\widetilde{K}_{NN} = K_{NN} - K_{NM}\left(K_{MM}\right)^{-1} K_{MN}. \label{eq:KNNtilde}
\end{align}
Here the matrices $ K $ are defined similarly as $ K_{NN} $ above, e.g.
$K_{NM} = \{k(x_n,x_m)\}_{n,m=1}^{N,M} $.

\subsection{Deep  Gaussian Processes}
\label{sec:mf}
A deep Gaussian Process (DGP)  is a hierarchical composition of GP models. We consider a model with $ L $ layers and $ T_l $ (stochastic) functions in layer $ l = 1,\dots,L $, i.e., a total number of $ T = \sum_{l=1}^{L} T_l $ functions~\cite{damianou2013dgp}. 
The input of layer $l$ is the output of the previous layer, $ f_N^l = [f^{l,1}(f_N^{l-1}), \dots,f^{l,T_l}(f_N^{l-1})] $, with starting values $f_N^{0} = x_N$.
We place independent GP priors augmented with inducing points on all the functions, using the same kernel $ k^l$ 
and the same set of inducing points $ x_M^l $ within layer $ l $. This leads to the following joint model density:
\begin{equation}\label{eq:dgpprior}
p(y_N,f_N,f_M) = p(y_N|f_N^L)\prod_{l=1}^{L}p(f_N^l|f_M^l;f_N^{l-1})p(f_M^l).
\end{equation}
Here $ p(f_M^l) = \prod_{t=1}^{T_l}\gauss{f_M^{l,t}}{0}{K^l_{MM}}$ and $ p(f_N^l|f_M^l;f_N^{l-1}) = \prod_{t=1}^{T_l}\gauss{f_N^{l,t}}{\widetilde{K}_{NM}^l f_M^{l,t}}{\widetilde{K}_{NN}^l} $,
where $ \widetilde{K}_{NM}^l$ and $\widetilde{K}_{NN}^l $ are given by the equivalents of Eq.~\eqref{eq:KNNtilde}, respectively.\footnote{\label{foot:mean}In order to avoid pathologies created by highly non-injective mappings in the DGP~\cite{duvenaud2014pathologies}, we follow Ref.~\cite{salimbeni-deisenroth-doubly-stochastic-vi-deep-gp} and add non-trainable linear mean terms given by the PCA mapping of the input data to the latent layers.
Those terms are omitted from the notation for better readability.}

Inference in this model~\eqref{eq:dgpprior} is intractable since we cannot marginalise over the latents $f^1_N, \ldots, f_N^{L-1}$ as they act as inputs to the non-linear kernel function.
We therefore choose to approximate the posterior by employing variational inference:
We search for an approximation $ q(f_N,f_M) $ to the true posterior $ p(f_N,f_M|y_N) $ by first choosing a variational family for the distribution $ q $ and then finding an optimal $ q $ within that family that minimises the Kullback-Leibler (KL) divergence $ \text{KL}[q||p] $. Equivalently, the so-called evidence lower bound (ELBO),
\begin{equation}\label{eq:elbogeneral}
\mathcal{L} = \int q(f_N,f_M)\log \frac{p(y_N,f_N,f_M)}{q(f_N,f_M)} df_N df_M,
\end{equation}
can be maximised.
In the following, we choose the variational family~\cite{salimbeni-deisenroth-doubly-stochastic-vi-deep-gp}
\begin{equation}\label{eq:dgpvarfam}
q(f_N,f_M) = q(f_M)\prod_{l=1}^{L}p(f_N^l|f_M^l;f_N^{l-1}).
\end{equation}
Note that $ f_M = \{f_M^{l,t} \}_{l,t=1}^{L, T_l}$ contains the inducing outputs of all layers, which might be covarying.
This observation will be the starting point for our structured approximation in Sec.~\ref{sec:fc}.

In the remaining part of this section, we follow Ref.~\cite{salimbeni-deisenroth-doubly-stochastic-vi-deep-gp} and restrict the distribution over the inducing outputs to be a-posteriori Gaussian and independent between different GPs (known as mean-field assumption, see also Fig.~\ref{fig:covariance}, left),
$
q(f_M) 
= \prod_{l=1}^L\prod_{t=1}^{T_l}q(f_M^{l,t}).
$
Here $q(f_M^{l,t}) = \gauss{f_M^{l,t}}{\mu_M^{l,t}}{S_M^{l,t}}$
and $ \mu_M^{l,t}$, $S_M^{l,t} $ are free variational parameters.
The inducing outputs $f_M$ act thereby as global latent variables that capture the information of the training data.
Plugging $q(f_M)$ into Eqs.~\eqref{eq:dgpprior},~\eqref{eq:elbogeneral},~\eqref{eq:dgpvarfam}, we can simplify the ELBO to
\begin{equation}\label{eq:dgpelbomf}
\mathcal{L}\!=\! \sum_{n=1}^{N}\!  \mathbb{E}_{q(f_n^L)} \! \left[\log p(y_n|f_n^L)\right]  - \! \sum_{l,t=1}^{L,T_l} \! \text{KL}[q(f_M^{l,t})||p(f_M^{l,t})].
\end{equation}
We first note that the ELBO decomposes over the data points, allowing for minibatch subsampling~\cite{hensman2013gaussian}.
However, the marginals of the output of the final layer, $ q(f_n^L) $, cannot be obtained analytically. 
While the mean-field assumption renders it easy to analytically marginalise out the inducing outputs (see Appx.~\ref{sec:appxdifMFFC}),
the outputs of the intermediate layers cannot be fully integrated out, since they are kernel inputs of the respective next layer, leaving us with
\begin{align}\label{eq:dgpsamplingmf}
q(f_n^L) = \int \prod_{l=1}^{L} q(f_n^l;f_n^{l-1})df_n^{1}\cdots df_n^{L-1},
&\ \text{where}
&
q(f_n^l;f_n^{l-1}) = \prod_{t=1}^{T_l} \gauss{f_n^{l,t}}{\widetilde{\mu}_n^{l,t}}{\widetilde{\Sigma}_n^{l,t}}.
\end{align}
The means and covariances are given by 
\begin{align}
{\widetilde{\mu}_n^{l,t}}   =  \widetilde{K}^l_{nM}\mu^{l,t}_M,
\quad
{\widetilde{\Sigma}_n^{l,t}} =  K^l_{nn} -
\widetilde{K}^l_{nM}\left(K^l_{MM} - S^{l,t}_M\right) \widetilde{K}^l_{Mn}.
\end{align}
We can straightforwardly obtain samples from $ q(f_n^L) $ by recursively sampling through the layers using Eq.~\eqref{eq:dgpsamplingmf}. 
Those samples can be used to evaluate the ELBO [Eq.~\eqref{eq:dgpelbomf}] and
to obtain unbiased gradients for parameter optimisation by using the reparameterisation trick~\cite{kingma2015variational,rezende2014stochastic}.
This stochastic estimator of the ELBO has low variance as we only need to sample over the local latent parameters
$f_n^1, \ldots, f_n^{L-1}$, while we can marginalise out the global latent parameters, i.e. inducing outputs, $f_M$.

%% file: subfiles/dgp.tex
\section{Structured Deep Gaussian Processes}
Next, we introduce a new class of variational families that allows to couple the inducing outputs $f_M$ within and across layers.
Surprisingly, analytical marginalisation over the inducing outputs $f_M$ is still possible after reformulating the problem into a recursive one that can be solved by induction. This enables an efficient inference scheme that refrains from sampling any global latent variables.
Our method generalises to arbitrary interactions which we exploit in the second part where we focus on the most prominent ones to attain speed-ups.
\subsection{Fully-Coupled DGPs}
\label{sec:fc}
We present now a new variational family that offers both, efficient computations and expressivity:
Our approach is efficient, since all global latent variables can be marginalised out 
, and expressive, since we allow for structure in the variational posterior.
We do this by leaving the Gaussianity assumption unchanged, while permitting dependencies between all inducing outputs (within layers and also across layers).
This corresponds to the (variational) ansatz
$
q(f_M) = \gauss{f_M}{\mu_M}{S_M}
$
with dimensionality $ TM $. 
By taking the dependencies between the latent processes into account, the resulting variational posterior $q(f_N,f_M)$  [Eq.~\eqref{eq:dgpvarfam}] is better suited to closely approximate the true posterior. 
 We give a comparison of exemplary covariance matrices $ S_M $ in Fig.~\ref{fig:covariance}.

Next, we investigate how the ELBO computations have to be adjusted when using the fully-coupled variational family.
Plugging  $q(f_M)$ 
into Eqs.~\eqref{eq:dgpprior},~\eqref{eq:elbogeneral} and~\eqref{eq:dgpvarfam}, yields
\begin{equation}\label{eq:elbodgpfc}
\mathcal{L}\! = \! \sum_{n=1}^{N} \! \mathbb{E}_{q(f_n^L)} \left[\log p(y_n|f_n^L)\right] - \text{KL}[q(f_M)||\prod_{l,t=1}^{L,T_l} p(f_M^{l,t})],
\end{equation}
which we derive in detail in Appx.~\ref{sec:appxelbo}.
The major difference to the mean-field DGP lies in the marginals $q(f_n^L)  $ of the outputs of the last layer:
Assuming (as in the mean-field DGP) that the distribution over the inducing outputs $ f_M $ factorises between the different GPs causes the marginalisation integral to factorise into $ L $ standard Gaussian integrals.
This is not the case for the fully-coupled DGP (see Appx.~\ref{sec:appxdifMFFC} for more details), which makes the computations more challenging.
The implications of using a fully coupled $ q(f_M) $ 
are summarised in the following theorem.
\begin{theorem} \label{theoremproof}
In a fully-coupled DGP as defined above, the marginals $q(f_n^L)$ can be written as
\begin{align}\label{eq:dgpsamplingfc}
q(f_n^L) = \int \prod_{l=1}^{L} q(f_n^l | f_n^1,\dots,f_n^{l-1})  df_n^{1}\cdots df_n^{L-1}
&  \ \text{where}
& q(f_n^l | f_n^1,\dots,f_n^{l-1}) = \gauss{f_n^l}{\hat{\mu}_n^l}{\hat{\Sigma}_n^l},
\end{align}
for each data point $ x_n $. The means and covariances are given by
\begin{align}
\label{eq:muhatn}
\hat{\mu}_n^l \! &= \! \widetilde{\mu}_n^l + \widetilde{S}_n^{l,1:l-1} \left( \widetilde{S}_n^{1:l-1,1:l-1} \right)^{-1} ( f_n^{1:l-1} - \widetilde{\mu}_n^{1:l-1}), 
\\
\label{eq:Sigmahatn}
\hat{\Sigma}_n^l \!  &= \! \widetilde{S}_n^{ll} - \widetilde{S}_n^{l,1:l-1} \left( \widetilde{S}_n^{1:l-1,1:l-1} \right)^{-1} \widetilde{S}_n^{1:l-1,l}, 
\end{align}
where  $\widetilde{\mu}_n^l = \widetilde{\mathcal{K}}^l_{nM} \mu_M^l$ and 
$\widetilde{S}_n^{ll'} = \delta_{ll'} \mathcal{K}^l_{nn} - \widetilde{\mathcal{K}}^l_{nM} \left(\delta_{ll'} \mathcal{K}^l_{MM} - S_M^{ll'}\right) \widetilde{\mathcal{K}}^{l'}_{Mn}$.
%
\end{theorem}

In Eqs.~\eqref{eq:muhatn} and \eqref{eq:Sigmahatn} the notation $ A^{l,1:l'} $ is used to index a submatrix of the variable $ A $, e.g.~${A}^{l,1:l'} = \left(A^{l,1} \cdots A^{l,l'} \right)$. 
Additionally, $ \mu_M^l  \in \mathbb{R}^{T_lM}$ denotes the subvector of $ \mu_M $ that contains the means of the inducing outputs in layer $ l $, and $ S_M^{ll'} \in \mathbb{R}^{T_l M \times T_{l'}M}$ contains the covariances between the inducing outputs of layers $ l $ and $ l' $.
For $\widetilde{\mu}_n^l$ and $\widetilde{S}_n^{ll'}$, we introduced the notation $ \mathcal{K}^l  = \left(\mathbb{I}_{T_l} \otimes K^l\right) $ as shorthand for the Kronecker product between the identity matrix $\mathbb{I}_{T_l}$ and the covariance matrix $ K^l $, and used $\delta$ for the Kronecker delta. 
{\new{
We verify in Appx.~\ref{sec:appxintuition2} that the formulas contain the mean-field solution as a special case by plugging in the respective covariance matrix.}}

By Thm.~\ref{theoremproof}, the inducing outputs $f_M$ can still be marginalised out, which enables low-variance estimators of the ELBO.
While the resulting formula for $q(f_n^l \vert f_n^1, \ldots, f_n^{l-1})$ has a similar form as Gaussian conditionals, this is only true at first glance (cf.~also Appx.~\ref{sec:appxintuition1}):
The latents of the preceding layers $f_n^{1:l-1}$ enter the mean $\hat{\mu}_n^l $ and the covariance matrix $\hat{\Sigma}_n^l $ also in an indirect way via  $\widetilde{S}_n$ as they appear as inputs to the kernel matrices.
\begin{proof}[\emph{\textbf{Sketch of the proof of Theorem \ref{theoremproof}}}]
	We start the proof with the general formula for $ q(f_n^L) $,
	\begin{equation}\label{eq:proofgen}
	q(f_n^L) = \int \left[ \int q(f_M)\prod_{l'=1}^L p(f_n^{l'} | f_M^{l'}) df_M \right] df_n^{1}\cdots df_n^{L-1},
	\end{equation}
	which is already (implicitly) used in Ref.~\cite{salimbeni-deisenroth-doubly-stochastic-vi-deep-gp} and which we derive in  Appx.~\ref{sec:appxfN}.
	In order to show the equivalence between the inner integral in Eq.~\eqref{eq:proofgen} and the integrand in Eq.~\eqref{eq:dgpsamplingfc} we proceed to find a recursive formula for integrating out the inducing outputs layer after layer:
	\begin{equation}\label{eq:proofform}
	\int q(f_M)\prod_{l'=1}^L p(f_n^{l'} | f_M^{l'}) df_M = \left[ \prod_{l'=1}^{l-1} q( f_n^{l'} | f_n^{1:l'-1})\right]
	\int q(f_n^l,f_M^{l+1:L} | f_n^{1:l-1}) \prod_{l'=l+1}^L p(f_n^{l'} | f_M^{l'}) df_M^{l'}.
	\end{equation}
	The equation above holds for $ l=1,\dots,L $ after the inducing outputs of layers $ 1,\dots,l $ have already been marginalised out.
	This is stated more formally in Lem.~\ref{lemmaind} in Appx.~\ref{sec:appxproof}, in which we also provide exact formulas for all terms.
Importantly, all of them are multivariate Gaussians with known mean and covariance.
	The lemma itself can be proved by induction 
	and we will show the general idea of the induction step here:
	For this, we assume the right hand side of Eq.~\eqref{eq:proofform} to hold for some layer $ l $ and then prove that it also holds for $ l\to l+1 $. We start by taking the (known) distribution within the integral and split it in two by conditioning on $ f_n^l $:
	\begin{eqnarray} \label{eq:proofstep1}
	q(f_n^l,f_M^{l+1:L} | f_n^{1:l-1}) = q(f_n^l | f_n^{1:l-1}) q(f_M^{l+1:L} | f_n^{1:l}) 
	\end{eqnarray}
	Then we show that the distribution $ q(f_n^l | f_n^{1:l-1}) $ can be written as part of the product in front of the integral in Eq.~\eqref{eq:proofform} (thereby increasing the upper limit of the product to $ l $). Next, we consider the integration over $ f_M^{l+1} $, where we collect all relevant terms (thereby increasing the lower limit of the product within the integral in Eq.~\eqref{eq:proofform} to $ l +2 $):
	\begin{align}
	&\int q(f_M^{l+1:L} | f_n^{1:l}) p(f_n^{l+1} | f_M^{l+1}) df_M^{l+1} 
	= \int q(f_M^{l+1} | f_n^{1:l}) q(f_M^{l+2:L} | f_n^{1:l}, f_M^{l+1}) p(f_n^{l+1} | f_M^{l+1}) df_M^{l+1} \nonumber \\
	\label{eq:prooflaststep}
	= &\int q(f_M^{l+1} | f_n^{1:l}) q(f_n^{l+1},f_M^{l+2:L} | f_n^{1:l}, f_M^{l+1}) df_M^{l+1} = q(f_n^{l+1},f_M^{l+2:L} | f_n^{1:l}).
	\end{align}
	The terms in the first line are given by Eqs.~\eqref{eq:proofstep1} and \eqref{eq:dgpprior}. 
All subsequent terms are also multivariate Gaussians that are obtained by standard operations like conditioning, joining two distributions, and marginalisation. 
We can therefore give an analytical expression of the final term in Eq.~\eqref{eq:prooflaststep}, which is exactly the term that is needed on the right hand side of Eq.~\eqref{eq:proofform} for $ l\to l+1 $. Confirming that this term has the correct mean and covariance completes the induction step.

After proving Lem.~\ref{lemmaind}, Eq.~\eqref{eq:proofform} can be used. For the case $ l=L $ the right hand side can be shown to yield $ \prod_{l=1}^{L} q(f_n^l | f_n^1,\dots,f_n^{l-1}) $. Hence, Eq.~\eqref{eq:dgpsamplingfc} follows by substituting the inner integral in Eq.~\eqref{eq:proofgen} by this term.
The full proof can be found in Appx.~\ref{sec:appxproof}.
\end{proof}
{\new{
Furthermore, we give a heuristic argument for Thm.~\ref{theoremproof} in Appx.~\ref{sec:appxintuition1} in which we show that by ignoring the recursive structure of the prior, the marginalisation of the inducing outputs $ f_M $ becomes straightforward. 
While mathematically not rigorous, the derivation provides additional intuition.}}

\begin{SCfigure}[3][t]
	\includegraphics[width=0.25 \textwidth]{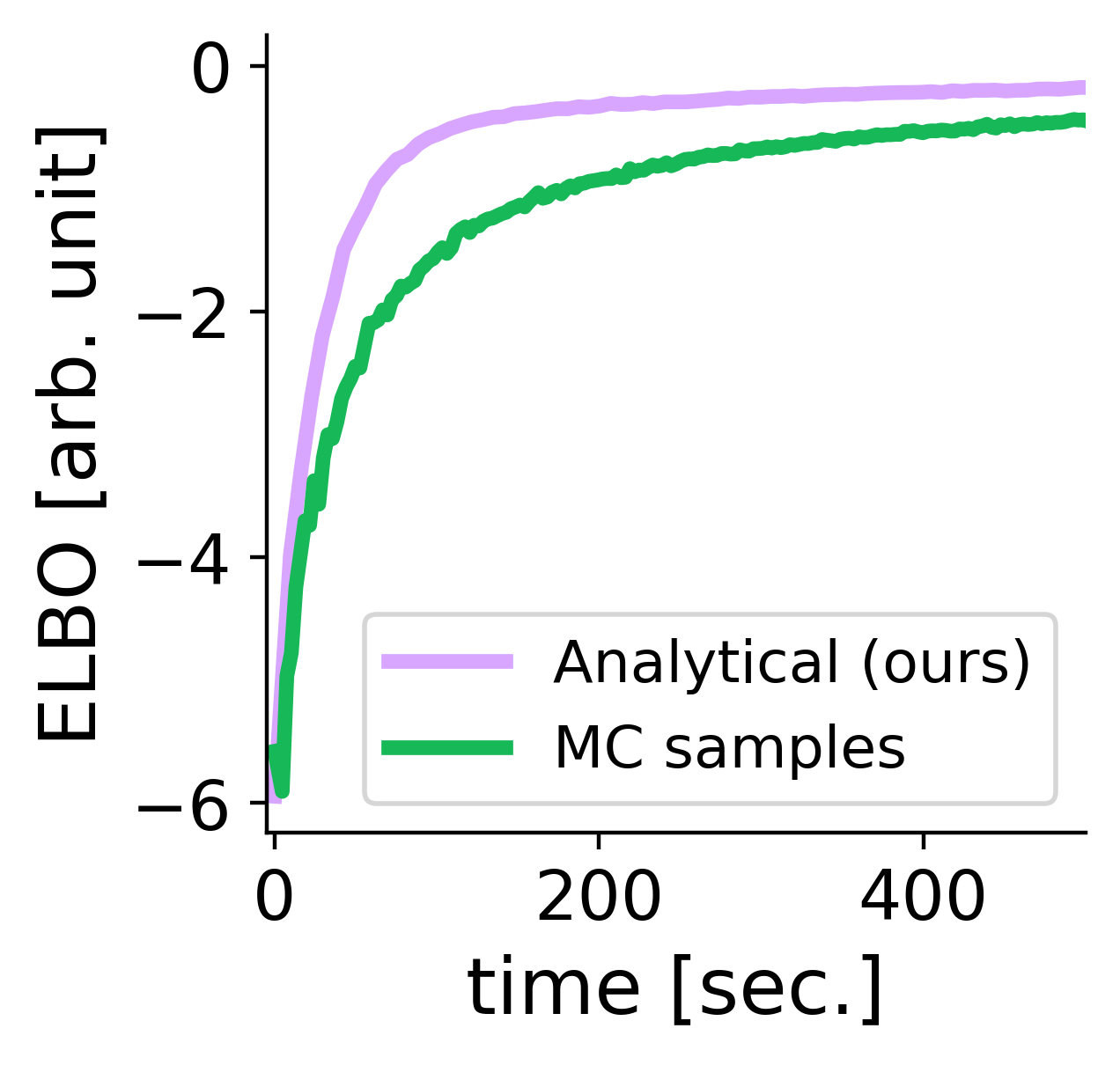}
	\caption{
		\textbf{Convergence behaviour: Analytical vs.~MC marginalisation.
		}
{\new{
		We plot the ELBO as a function of time in seconds when the marginalisation of the inducing outputs $ f_M $ is performed analytically via our Thm.~\ref{theoremproof} (purple)
and via MC sampling (green).
We used a fully-coupled DGP with our standard three layer architecture (see Sec.~\ref{sec:afc}), on the \textit{concrete} UCI dataset trained with Adam \cite{kingma2014adam}.}}}
	\label{fig:convergence}
\end{SCfigure}
Next, we use our novel variational approach to fit a fully coupled DGP model with $L=3$ layers to the \textit{concrete} UCI dataset.
We can clearly observe that this algorithmic work pays off: 
Fig.~\ref{fig:covariance} shows that there is more structure in the covariance matrix $ S_M $ than the mean-field approximation allows.
This additional structure results in a better approximation of the true posterior as we validate on a range of benchmark datasets (see Tab.~\ref{tab:elbo-mf-fc} in Appx.~\ref{sec:appxexdet}) for which we observe larger ELBO values for the fully-coupled DGP than for the mean-field DGP.
{\new{
Additionally, we show in Fig.~\ref{fig:convergence} that our analytical marginalisation over the inducing outputs $ f_M $ leads to faster convergence compared to Monte Carlo (MC) sampling, since the corresponding ELBO estimates have lower variance.
Independently from our work, the sampling-based approach has also been proposed in Ref.~\cite{ustyuzhaninov2019compositional}.
}}

However, in comparison with the mean-field DGP, the increase in the number of variational parameters also leads to an increase in runtime and made convergence with standard optimisers fragile due to many local optima.
We were able to circumvent the latter by the use of natural gradients~\cite{amari1998natgrads}, which have been found to work well for (D)GP models before~\cite{hebbal2019bayesian,salimbeni2018natgrads,salimbeni2019deep}, 
but this increases the runtime even further (see Sec.~\ref{sec:runtime}).
It is therefore necessary
to find a smaller variational family if we want to use the method in large-scale applications.

An optimal variational family combines the best of both worlds, i.e., being as efficient as the mean-field DGP while retaining the most important interactions introduced in the fully-coupled DGP.
We want to emphasise that there are many possible ways of restricting the covariance matrix $S_M$ that potentially lead to benefits in different applications.
For example, the recent work~\cite{ustyuzhaninov2019compositional} studies the compositional uncertainty in deep GPs using a particular restriction of the inverse covariance matrix.
The authors also provide specialised algorithms to marginalise out the inducing outputs in their model.
Here, we provide an analytic marginalisation scheme for arbitrarily structured covariance matrices that will vastly simplify future development of application-specific covariances.
Through the general framework that we have developed, testing them is straightforward and can be done via simply implementing a naive version of the covariance matrix in our code.\footnote{\new{Python code (building on code for the mean-field DGP~\cite{salimbeni2019deep},  GPflow~\cite{matthews2017gpflow} and TensorFlow~\cite{abadi2016tensorflow}) implementing our method is provided at \url{https://github.com/boschresearch/Structured_DGP}.}
A pseudocode description of our algorithm is given in Appx.~\ref{sec:pseudocode}.}
In the following, we propose one possible class of covariance matrices based on our empirical findings. 

\subsection{Stripes-and-Arrow Approximation}
\label{sec:afc}
In this section, we describe a new variational family that trades off efficiency and expressivity by sparsifying the covariance matrix $S_M$.
Inspecting Fig.~\ref{fig:covariance} (right) again, 
we observe 
  that besides the $ M\times M $ blocks on the diagonal, the diagonal stripes~\cite{smolarski2006diagonally-striped} (covariances between the GPs in latent layers at the same relative position), and an arrow structure (covariances from every intermediate layer GP to the output GP) receive large values.
We make similar observations also for different datasets and different DGP architectures as shown in Fig.~\ref{fig:add_cov} in Appx.~\ref{sec:appxexdet}.
{\new{Note that the stripes pattern can also be motivated theoretically as we expect the residual connections realised by the mean functions (footnote \ref{foot:mean}) to lead to a coupling between successive latent GPs.}}
We therefore propose as one special form to keep only these terms and neglect all other dependencies by setting them to zero in the covariance matrix, resulting in a structure consisting of an arrowhead and diagonal stripes (see Fig.~\ref{fig:covariance} middle).

Denoting the number of GPs per latent layer as $ \tau $, it is straightforward to show that the number of non-zero elements in the covariance matrices of mean-field DGP, stripes-and-arrow DGP, and fully-coupled DGP scale as $ \mathcal{O}(\tau L M^2) $, $ \mathcal{O}(\tau L^2 M^2) $, and $ \mathcal{O}(\tau^2 L^2 M^2) $, respectively.
In the example of Fig.~\ref{fig:covariance}, we have used $ \tau = 5 $, $ L=3 $, and $ M=128 $, yielding $ 1.8\times 10^5 $, $ 5.1\times 10^5 $, and $ 2.0\times 10^6 $ non-zero elements in the covariance matrices.
Reducing the number of parameters already leads to shorter training times since less gradients need to be computed. Furthermore, the property that makes this form so compelling is that the covariance matrix $ \widetilde{S}_n^{1:l-1,1:l-1} $ [needed in Eqs.~\eqref{eq:muhatn} and \eqref{eq:Sigmahatn}]  as well as
the Cholesky decomposition\footnote{In order to ensure that $ S_M $ is positive definite, we will numerically exclusively work with its Cholesky factor $ L $, a unique lower triangular matrix such that $ S_M = LL^\top $.} of $ S_M $ have the same sparsity pattern.
Therefore only the non-zero elements at pre-defined positions have to be calculated which is explained in Appx.~\ref{sec:appxLA}. 
The complexity for the ELBO is $ \mathcal{O}(NM^2\tau L^2 + N \tau^3 L^3 + M^3\tau L^3) $. This is a moderate increase compared to the mean-field DGP whose ELBO has complexity $ \mathcal{O}(NM^2\tau L) $, while it is a clear improvement over the fully-coupled approach with complexity $ \mathcal{O}(NM^2\tau^2 L^2 + N \tau^3L^3 +M^3\tau^3L^3) $ (see Appx.~\ref{sec:appxLA} for derivations).
An empirical runtime comparison is provided in Sec.~\ref{sec:runtime}.

After having discussed the advantages of the proposed approximation a remark on a disadvantage is in order:
The efficient implementation of Ref.~\cite{salimbeni2018natgrads} for natural gradients cannot be used in this setting, since the transformation from our parameterisation to a fully-parameterised multivariate Gaussian is not invertible. 
However, this is only a slight disadvantage 
since the stripes-and-arrow approximation has a drastically reduced number of parameters, compared to the fully-coupled approach, and we experimentally do not observe the same convergence problems when using standard optimisers (see Appx.~\ref{sec:appxexdet}, Fig.~\ref{fig:elbo}).

%% file: subfiles/related_work.tex
\subsection{Joint sampling of global and local latent variables}
In contrast to our work, Refs.~\cite{havasi2018inference, yu2019implicit} drop the Gaussian assumption over the inducing outputs $f_M$ and allow instead for potentially multi-modal approximate posteriors.
While their approaches are arguably more expressive than ours, their flexibility comes at a price:
the distribution over the inducing outputs $f_M$ is only given implicitly in form of Monte Carlo samples.
Since the inducing outputs $f_M$ act as global latent parameters, the noise attached to their sampling-based estimates affects all samples from one mini-batch. This can often lead to higher variances which may translate to slower convergence~\cite{kingma2015variational}.
We compare to Ref.~\cite{havasi2018inference} in our experiments.

%% file: subfiles/experiments.tex
\section{Experiments}
\label{sec:experiments}
In Sec.~\ref{sec:benchmark}, we study the predictive performance of our stripes-and-arrow approximation.
Since it is difficult to assess accuracy and calibration on the same task, we ran a joint study of interpolation and extrapolation tasks, where in the latter the test points are distant from the training points.
We found that the proposed approach balances accuracy and calibration, thereby outperforming its competitors on the combined task.
Examining the results for the extrapolation task more closely, we find that our proposed method significantly outperforms the competing DGP approaches.
In Sec.~\ref{sec:runtime}, we assess the runtime of our methods and confirm that our approximation has only a negligible overhead compared to mean-field and is more efficient than a fully-coupled DGP.
Due to space constraints, we moved many of the experimental details to Appx.~\ref{sec:appxexdet}.

\subsection{Benchmark Results}
\label{sec:benchmark}
We compared the predictive performance of our efficient \underline{st}ripes-and-\underline{ar}row approximation (STAR DGP) with a mean-field approximation (MF DGP)~\cite{salimbeni-deisenroth-doubly-stochastic-vi-deep-gp}, stochastic gradient Hamiltonian Monte Carlo (SGHMC DGP)~\cite{havasi2018inference} and a sparse GP (SGP)~\cite{hensman2013gaussian}.
As done in prior work, we report results on eight UCI datasets and employ as evaluation criterion the average marginal test log-likelihood (tll).

\begin{table*}[t!]
	\caption{
		\textbf{Interpolation behaviour on UCI benchmark datasets.}
		We report marginal tlls (the larger, the better) for various methods, where $ L $ denotes the number of layers. 
		Standard errors are obtained by repeating the experiment 10 times.
		We marked all methods in bold that performed better or as good as the standard sparse GP.
	}
	\setlength{\tabcolsep}{4pt}
	\begin{scriptsize}
		\begin{center}
			\begin{tabular}{l|c|ccc|cc|cc}
				\toprule
				Dataset & SGP & 
				\multicolumn{3}{c}{SGHMC DGP} &  
				\multicolumn{2}{c}{MF DGP} &
				\multicolumn{2}{c}{STAR DGP}  
				\\
				(N,D) & L1 & L1 & L2 & L3 &   L2 & L3 & L2 & L3
				\\
				\midrule
				boston (506,13) & -2.58(0.10) & -2.75(0.18) &\textbf{ -2.51(0.07) }& \textbf{-2.53(0.09)} &\textbf{ -2.43(0.05)} & \textbf{-2.48(0.06)} & \textbf{-2.47(0.08)} &\textbf{ -2.43(0.05)}\\
				energy (768, 8) & -0.71(0.03) & -1.16(0.44) & \textbf{-0.37(0.12)} &\textbf{ -0.34(0.11)} & \textbf{-0.73(0.02)} & -0.75(0.02) & -0.75(0.02) & -0.75(0.02)\\
				concrete (1030, 8) & -3.09(0.02) & -3.50(0.34) &\textbf{ -2.89(0.06)} &\textbf{ -2.88(0.06)} &\textbf{ -3.06(0.03)} & \textbf{-3.09(0.02)} &\textbf{ -3.04(0.02)} &\textbf{ -3.05(0.02)}\\
				wine red (1599,11) & -0.88(0.01) &-0.90(0.03) & \textbf{-0.81(0.03)} & \textbf{-0.80(0.07)} & { -0.89(0.01)} &{ -0.89(0.01)} &\textbf{ -0.88(0.01)} & \textbf{-0.88(0.01)}\\
				kin8nm (8192, 8) & 1.05(0.01) & \textbf{1.14(0.01) }& \textbf{1.38(0.01)} &\textbf{ 1.25(0.14)} & \textbf{1.30(0.01)} &\textbf{ 1.31(0.01)} & \textbf{1.28(0.01)} & \textbf{1.29(0.01)}\\
				power (9568, 4) & {-2.78(0.01)} & \textbf{-2.75(0.02)} &\textbf{ -2.68(0.02)} & \textbf{-2.65(0.02)} & \textbf{-2.77(0.01)} & \textbf{-2.76(0.01)} &\textbf{ -2.77(0.01)} &\textbf{ -2.77(0.01)}\\
				naval (11934,16) & 7.56(0.09) &\textbf{ 7.77(0.04)} & 7.32(0.02) & 6.89(0.43) & 7.11(0.11) & 7.05(0.09) & 7.06(0.08) & 6.25(0.31)\\
				protein (45730, 9) & -2.91(0.00) &\textbf{ -2.76(0.00)} &\textbf{ -2.64(0.01)} & \textbf{-2.58(0.01)} &\textbf{ -2.83(0.00)} &\textbf{ -2.79(0.00)} &\textbf{ -2.83(0.00)} & \textbf{-2.80(0.00)}\\
				\bottomrule
			\end{tabular}
		\end{center}
		\label{tab:interpolation}
	\end{scriptsize}
\end{table*}

We assessed the interpolation behaviour of the different approaches by randomly partitioning the data into a training and a test set  with a $90:10$ split.
To investigate the extrapolation behaviour, we created test instances that are distant from the training samples:
We first randomly projected the inputs $ X $ onto a one-dimensional subspace $z = Xw$, where the weights $w \in \mathbb{R}^D$ were drawn from a standard Gaussian distribution. 
We subsequently ordered the samples w.r.t.~$z$ and divided them accordingly into training and test set using a $50:50$ split.

We first confirmed the reports from the literature~\cite{havasi2018inference,salimbeni-deisenroth-doubly-stochastic-vi-deep-gp}, that DGPs have on interpolation tasks an improved performance compared to sparse GPs (Tab.~\ref{tab:interpolation}). 
{\new{We also observed that in this setting SGHMC outperforms the MF DGP and our method, which are on par.

Subsequently, we performed the same analysis on the extrapolation task.
While our approach, STAR DGP, seems to perform slightly better than MF DGP and also SGHMC DGP, the large standard errors of all methods hamper a direct comparison (see Tab.~\ref{tab:extrapolation} in Appx.~\ref{sec:appxexdet}).
This is mainly due to the random 1D-projection of the extrapolation experiment: The direction of the projection has a large impact on the difficulty of the prediction task. Since this direction changes over the repetitions, the corresponding test log-likelihoods vary considerably, leading to large standard errors.

We resolved this issue by performing a direct comparison between STAR DGP and the other two DGP variants:
To do so, we computed the frequency of test samples for which STAR DGP obtained a larger log-likelihood than MF/SGHMC DGP on each train-test split independently. Average frequency $ \mu $ and its standard error $ \sigma $ were subsequently computed over 10 repetitions and are reported in Tab.~\ref{tab:DGP-test}. On 5/8 datasets STAR DGP significantly outperforms MF DGP and SGHMC DGP ($ \mu > 0.50 + \sigma $), respectively, while the opposite only occurred on \textit{kin8nm}. 
In Tab.~\ref{tab:DGP-test-add} in Appx.~\ref{sec:appxexdet}, we show more comparisons, that also take the absolute differences in test log likelihoods into account and additionally consider the comparison of fully-coupled and MF DGP.
Taken together, we conclude that our structured approximations are in particular beneficial in the extrapolation scenario, while their performance is similar to MF DGP in the interpolation scenario.
}}

\begin{SCtable}[2][t]
	\caption{
		{\new{\textbf{Extrapolation behaviour: direct comparison of DGP methods.}
		Average frequency $ \mu $ and its standard error $ \sigma $ (computed over 10 repetitions) of the STAR DGP outperforming the MF DGP (left) and the SGHMC DGP (right) on the marginal tll of individual repetitions of the extrapolation task (see main text for details). Results are for DGPs with three layers. We mark numbers in bold (italics) if STAR outperforms its competitor (vice versa).}}
	}
	\setlength{\tabcolsep}{4pt}
	\begin{scriptsize}
			\begin{tabular}{ l  | c c }
				\toprule
				Dataset &    \multicolumn{1}{c}{MF vs.~STAR} & \multicolumn{1}{c}{SGHMC vs.~STAR}\\
				\midrule
				boston&   \textbf{0.55(0.04)}    &  0.50(0.05)    \\
				energy  & \textbf{0.73(0.05)}  & \textbf{0.60(0.04)}       \\
				concrete  &    \textbf{0.57(0.04)}    & \textbf{0.60(0.03)} \\
				wine red    &   \textbf{0.57(0.04)}& \textbf{0.63(0.02)}   \\
				kin8nm  & \textit{0.36(0.03)}   &  \textit{0.44(0.05)}   \\
				power  &  {0.44(0.06)}  & \textbf{0.64(0.03)}    \\
				naval  &  \textbf{0.67(0.06)}&  \textbf{0.58(0.03)}  \\
				protein  &  0.49(0.03)   & 0.50(0.03) \\
				\bottomrule
			\end{tabular}
		\label{tab:DGP-test}
	\end{scriptsize}
\end{SCtable}

\begin{SCfigure}[1.2][t]
	\caption{
		\textbf{Calibration Study.}
		Left: While the predicted variances increase for all methods as a function of the distance to the training data, we find that at any given distance, the uncertainty decreases from SGP to STAR DGP to MF DGP. 
		Right: 
		We plot the mean squared error as a function of the predicted variance.
		If the mean squared error is larger than the predicted variance, the latter underestimates the uncertainty.
		Results are recorded on the \textit{kin8nm} UCI dataset and smoothed for plotting by using a median filter.
	}
	\label{fig:zoom}
	\includegraphics[width=0.45 \columnwidth]{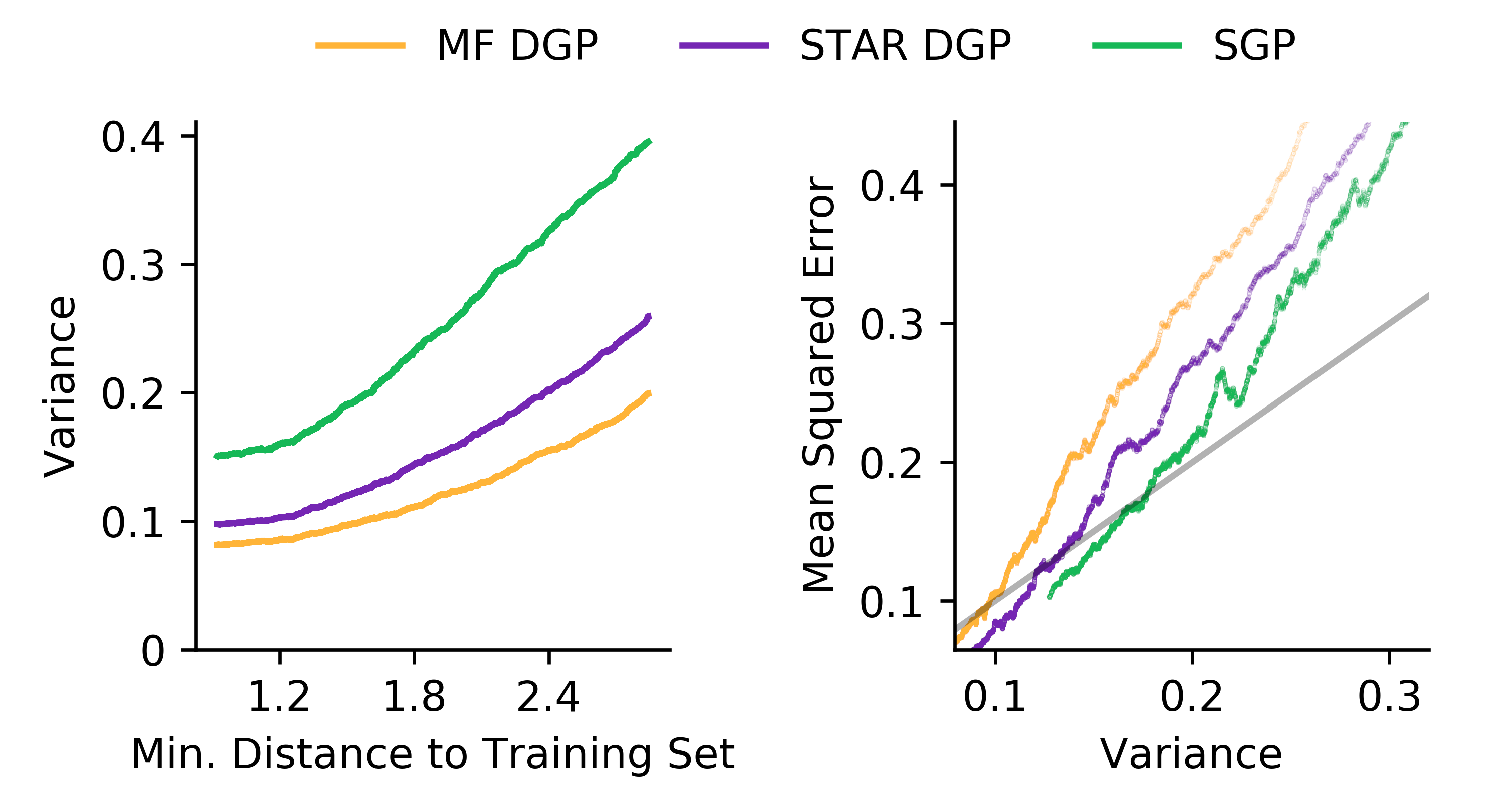}
\end{SCfigure}
{\new{Next, we performed an in-depth comparison between the approaches that analytically marginalise the inducing outputs:}}
In Fig.~\ref{fig:zoom} we show  that the predicted variance ${ \sigma}_*^2$ increased as we moved away from the training data (left) while the mean squared errors also grew with larger ${ \sigma}_*^2$ (right).
The mean squared error is an empirical unbiased estimator of the variance  $\text{Var}_* = \mathbb{E}[(y_* - \mu_*)^2]$ where $y_*$ is the test output and $\mu_*$ the mean predictor.
The predicted variance ${ \sigma}^2_*$ is also an estimator of $\text{Var}_*$.
It is only unbiased if the method is calibrated.
However, we observed for the mean-field approach that,
when moving away from the training data,  the mean squared error was larger than the predicted variances pointing towards underestimated uncertainties. 
While the mean squared error for SGP matched well with the predictive variances, the predictions are rather inaccurate as demonstrated by the large predicted variances.
Our method reaches a good balance, having generally more accurate mean predictions than SGP and at the same time more accurate variance predictions than MF DGP.

Finally, we investigated the behaviour of the SGHMC approaches in more detail.
We first ran a one-layer model that is equivalent to a sparse GP but with a different inference scheme:
Instead of marginalising out the inducing outputs, they are sampled.
We observed that the distribution over the inducing outputs is non-Gaussian (see Appx.~\ref{sec:appxexdet}, Fig.~\ref{fig:sghmc}), even though the optimal approximate posterior distribution is provably Gaussian in this case~\cite{titsias2009variational}.
A possible explanation for this are convergence problems since the global latent variables are not marginalised out, which, in turn, offers a potential explanation for the poor extrapolation behaviour of SGHMC that we observed in our experiments across different architectures and datasets. Similar convergence problems have also been observed by Ref.~\cite{salimbeni2019deep}.

\subsection{Runtime}
\label{sec:runtime}

\begin{SCfigure}[3]
\caption{
\textbf{Runtime comparison.}
We compare the runtime of our efficient STAR DGP versus the FC DGP and the 
MF DGP on the \textit{protein} UCI dataset. 
Shown is the runtime of one gradient step in seconds on a logarithmic scale as a function of the number of inducing points $M$. 
The dotted grey lines show the theoretical runtime $\mathcal{O}(M^2) $.}
		\includegraphics[width=0.3 \textwidth]{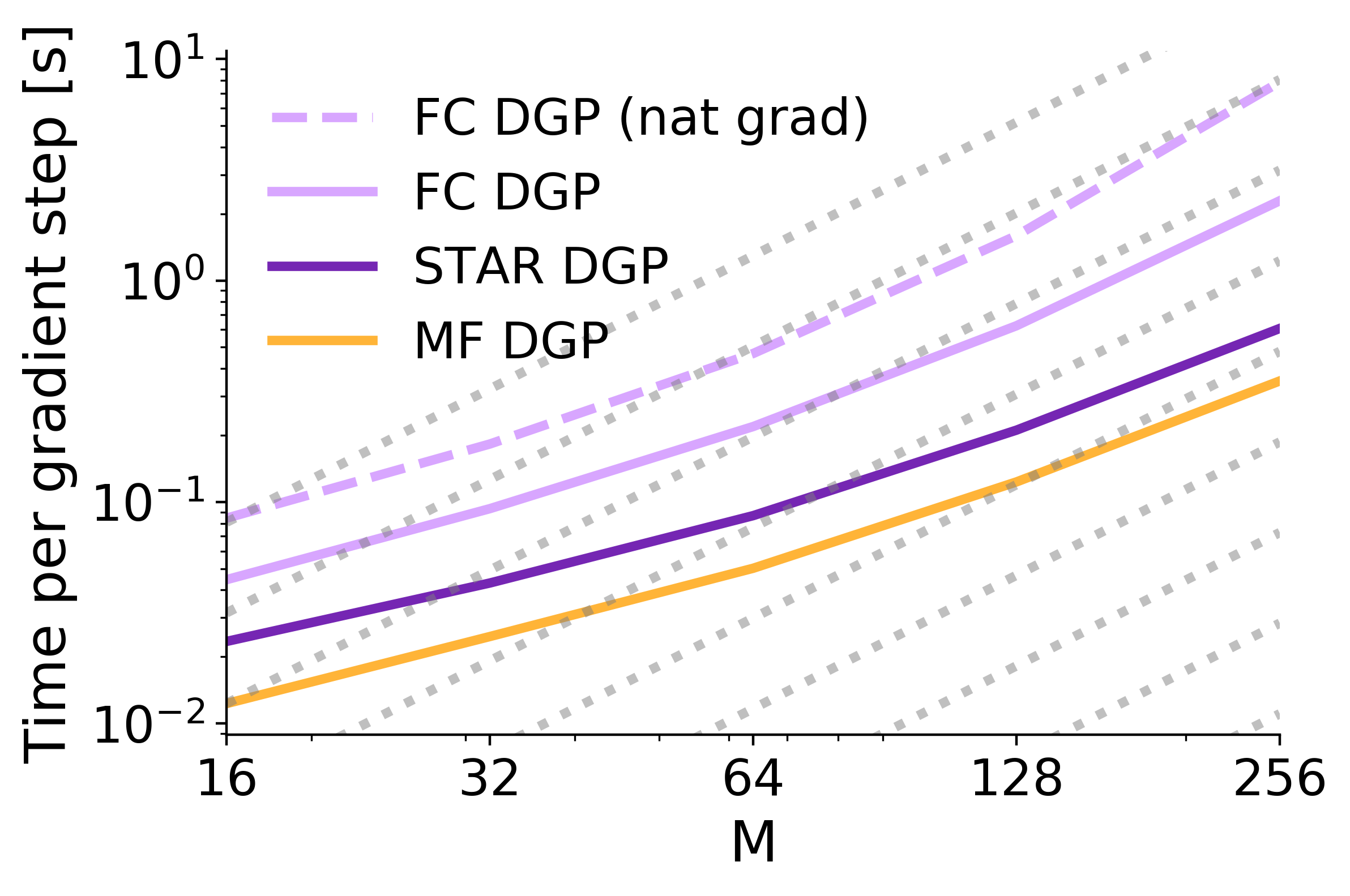}
\label{fig:runtime}
\end{SCfigure}

We compared the impact of the variational family on the runtime as a function of the number of inducing points $M$.
For the fully-coupled (FC) variational model, we also recorded the runtime when employing natural gradients~\cite{salimbeni2018natgrads}.
The results can be seen in Fig.~\ref{fig:runtime}, where the order from fastest to slowest method was
proportional to the complexity of the variational family: mean-field, stripes-and-arrow, fully-coupled DGP.
For our standard setting, $M=128$,  our STAR approximation was only two times slower than the mean-field but three times faster than FC DGP (trained with Adam~\cite{kingma2014adam}).
This ratio stayed almost constant when the number of inducing outputs $ M $ was changed, since the most important term in the computational costs scales as $ \mathcal{O}(M^2) $ for all methods. 
Subsequently, we performed additional experiments in which we varied the architecture parameters $L$ and $\tau$. 
Both confirm that the empirical runtime performance scales with the complexity of the variational family (see Appx.~\ref{sec:appxexdet}, Fig.~\ref{fig:runtime2}) and matches our theoretical estimates in Sec.~\ref{sec:afc}.

%% file: subfiles/summary.tex
\section{Summary}
In this paper, we investigated a new class of variational families for deep Gaussian processes (GPs).
Our approach is (i) efficient as it allows to marginalise analytically over the global latent variables and 
(ii) expressive as it couples the inducing outputs across layers in the variational posterior.
Naively coupling all inducing outputs does not scale to large datasets, hence we suggest a sparse and structured approximation that only takes the most important dependencies into account.
In a joint study of interpolation and extrapolation tasks as well as in a careful evaluation of the extrapolation task on its own, our approach outperforms its competitors, since it balances accurate predictions and calibrated uncertainty estimates.
{\new{
Further research is required to understand why our structured approximations are especially helpful for the extrapolation task. One promising direction could be to look at differences of inner layer outputs (as done in Ref.~\cite{ustyuzhaninov2019compositional}) and link them to the final deep GP outputs.
}}

There has been a lot of follow-up work on deep GPs in which the probabilistic model is altered to allow for multiple outputs~\cite{kaiser2018bayesian}, multiple input sources~\cite{hamelijnck2019multi}, latent features~\cite{salimbeni2019deep} or for interpreting the latent states as differential flows~\cite{hegde2018deep}.
Our approach can be easily adapted to any of these models and is therefore a promising line of work to advance inference in deep GP models.

Our proposed structural approximation is only one way of coupling the latent GPs.
Discovering new variational families that allow for more speed-ups either by 
applying Kronecker factorisations as done in the context of neural networks~\cite{martens2015optimizing}, placing a grid structure over the inducing inputs~\cite{izmailov2017scalable}, or by taking a conjugate gradient perspective on the objective~\cite{wang2019exact} are interesting directions for future research.
{\new{Furthermore, we think that the dependence of the optimal structural approximation on various factors (model architecture, data properties, etc.) is worthwhile to be studied in more detail.}}

%% file: subfiles/impact.tex
\section*{Broader Impact}
In many applications, machine learning algorithms have been shown to achieve superior predictive performance compared to hand-crafted or expert solutions~\cite{silver2016mastering}.
However, these methods can be applied in safety-critical applications only if they return predictive distributions allowing to quantify the uncertainty of the prediction~\cite{leibig2017leveraging}.
For instance, a medical diagnosis tool can be applied only if each diagnosis is endowed with a confidence interval such that in case of ambiguity a physician can be contacted.
Our work yields accurate predictive distributions for deep non-parametric models by allowing correlations between and across layers in the variational posterior. 
As we validate in our experiments, this also holds true when the input distribution at test time differs from the input distribution at training time.
In our medical example, this might be the case if the hospital where the data is recorded is different from the one where the diagnosis tool is deployed.

\section*{Acknowledgements}
{\new{We thank Buote Xu for valuable comments and suggestions on an early draft of the paper. We furthermore acknowledge the detailed and constructive feedback from the four anonymous reviewers, particularly for suggesting a new experiment which lead to Fig.~\ref{fig:convergence}.}}
\bibliography{main}
\bibliographystyle{plain}

%% file: subfiles/appendix_theory.tex
\clearpage
\newgeometry{
	textheight=8.5in,
	textwidth=6.5in,
	top=1in,
	headheight=12pt,
	headsep=25pt,
	footskip=30pt
}

\appendix


\begin{center}
{\Large{Supplementary material for}}

\bigskip

{\LARGE{\textbf{Beyond the Mean-Field:}}}

\smallskip

{\LARGE{\textbf{Structured Deep Gaussian Processes Improve the Predictive Uncertainties}}}

\end{center}
\section{Marginalisation of the inducing outputs (proof of Theorem 1)} \label{sec:appxproof}

The aim of this section is to provide a complete proof for Thm.~\ref{theoremproof}.
We will do this by starting from the formula for $ q(f_n^l) $ that we work out in Appx.~\ref{sec:appxfN},
\begin{equation}\label{eq:appxqfnLproof0}
q(f_n^L) = \int \left[ \int q(f_M)\prod_{l=1}^L p(f_n^l | f_M^l; f_n^{l-1}) df_M \right] df_n^{1}\cdots df_n^{L-1}.
\end{equation}
Comparing to Eq.~\eqref{eq:dgpsamplingfc}, we see that it remains to be shown that indeed
\begin{equation}\label{eq:appxmain}
\int q(f_M)\prod_{l=1}^L p(f_n^l | f_M^l; f_n^{l-1}) df_M = \prod_{l=1}^{L} q(f_n^l | f_n^1,\dots,f_n^{l-1}),
\end{equation}
where the distributions $ q $ on the right hand side have the properties described in Eqs.~\eqref{eq:dgpsamplingfc} - \eqref{eq:Sigmahatn}.
The terms appearing on the left hand side are given by $ q(f_M) = \gauss{f_M}{\mu_M}{S_M} $, which is interchangeably also denoted as
\begin{equation}\label{eq:appxqfM}
q(f_M^{1:L}) = \gauss{f_M^{1:L}}{\mu_M^{1:L}}{S_M^{1:L,1:L}} = q
\begin{pmatrix}
f_M^1\\
\vdots\\
f_M^L
\end{pmatrix} = \mathcal{N}
\left(
\begin{pmatrix}
	f_M^1\\
	\vdots\\
	f_M^L
\end{pmatrix}
\middle|
\begin{pmatrix}
	\mu_M^1\\
	\vdots\\
	\mu_M^L
\end{pmatrix},
\begin{pmatrix}
	S_M^{11} & \cdots & S_M^{1L}\\
	\vdots & \ddots & \vdots\\
	S_M^{L1} & \cdots & S_M^{LL}
\end{pmatrix}
\right),
\end{equation}
and
\begin{equation}\label{eq:appxpcond}
p(f_n^l | f_M^l; f_n^{l-1}) = \gauss{ f_n^l }{ \widetilde{\mathcal{K}}_{nM}^l f_M^l}{ \widetilde{\mathcal{K}}_{nn}^l },
\end{equation}
where
\begin{alignat}{2}\label{eq:appxKnM}
&\widetilde{\mathcal{K}}_{nM}^l &&= \mathcal{K}_{nM}^l \left( \mathcal{K}_{MM}^l \right)^{-1} \\ \label{eq:appxKnn}
&\widetilde{\mathcal{K}}_{nn}^l &&= \mathcal{K}_{nn}^l - \mathcal{K}_{nM}^l \left( \mathcal{K}_{MM}^l \right)^{-1} \mathcal{K}_{Mn}^l.
\end{alignat}

In order to show that Eq.~\eqref{eq:appxmain} holds, we will introduce a rather technical lemma in the following and prove it later by induction.

\begin{lemma}
	\label{lemmaind}
	Given the definitions in Eqs.~\eqref{eq:appxqfM} and \eqref{eq:appxpcond}, $\forall l=1,\dots,L$ we have
	\begin{equation}
	\label{eq:indmain}
	\int q(f_M^{1:L})\prod_{l'=1}^L p(f_n^{l'} | f_M^{l'}) df_M^{l'} =
	\left[ \prod_{l'=1}^{l-1} q( f_n^{l'} | f_n^{1:l'-1})\right]
	\int q(f_n^l,f_M^{l+1:L} | f_n^{1:l-1}) \prod_{l'=l+1}^L p(f_n^{l'} | f_M^{l'}) df_M^{l'},
	\end{equation}
	where $ q( f_n^{l'} | f_n^{1:l'-1}) $ is as in Eq.~\eqref{eq:dgpsamplingfc} and
	
	\begin{equation}\label{eq:indgauss}
	q(f_n^l,f_M^{l+1:L}| f_n^{1:l-1}) = \gauss{\begin{pmatrix*}[l]
		f_n^l \\ f_M^{l+1:L}
		\end{pmatrix*}}{\begin{pmatrix*}[l]
		\phantom{^l}\hat{\mu}_n^l \\ ^l\hat{\mu}_M^{l+1:L}
		\end{pmatrix*}}{\begin{pmatrix*}
		\hat{\Sigma}^l_n & ^l\hat{\Sigma}_{nM}^{l,l+1:L} \\
		\left(^l\hat{\Sigma}_{nM}^{l,l+1:L}\right)^\top & ^l\hat{\Sigma}_{M}^{l+1:L,l+1:L}
		\end{pmatrix*}}.
	\end{equation}
	Here $ \hat{\mu}_n^l $ and $ \hat{\Sigma}^l_n $ are as in Eqs.~\eqref{eq:muhatn} and \eqref{eq:Sigmahatn}, respectively, and we defined
	\begin{equation}\label{eq:indhelp}
	\begin{aligned}
	^l\hat{\mu}_M^{l+1:L} &= \mu_M^{l+1:L} + \SM{l+1:L}{1:l-1} \lemdiagK{Mn}{l-1} \Sninv ( f_n^{1:l-1} - \widetilde{\mu}_n^{1:l-1}) \\
	^l\hat{\Sigma}_{M}^{l+1:L,l+1:L} &= \SM{l+1:L}{l+1:L} - \SM{l+1:L}{1:l-1}
	\lemdiagK{Mn}{l-1} \Sninv \lemdiagK{nM}{l-1} \SM{1:l-1}{l+1:L}\\
	^l\hat{\Sigma}_{nM}^{l,l+1:L} &= \widetilde{\mathcal{K}}_{nM}^l \SM{l}{l+1:L} -	\Sn{l}{1:l-1} \Sninv \lemdiagK{nM}{l-1}\SM{1:l-1}{l+1:L}.
	\end{aligned}
	\end{equation}
\end{lemma}

In the equations above we used $ \text{diag}(A^{1:l}) $ to denote the formation of a block diagonal matrix, where the diagonal blocks are given by $ A^1,\dots,A^l $.
Note that while we only need one index to label $ \hat{\mu}_n^l $ and $ \hat{\Sigma}^l_n $, we need several for the objects defined in Eq.~\eqref{eq:indhelp}.
Take e.g.~$^l\hat{\Sigma}_{M}^{l+1:L,l+1:L}$: The upper left index denotes for which $ l $ the formula is valid (which will become important when we do the induction step $ l \to l+1 $).
The upper right indices (try to) capture which terms of $ S_M $ are most important for the definition, they have nothing to do with the dimensionality of the objects.
(In fact, the matrix $ ^l\hat{\Sigma}_{M} $ contains $ L-l-1 \times L-l-1 $ blocks of various sizes $ T_lM \times T_{l'}M $.)
This makes it easier later on when we do calculations with these objects.

Before we prove Lem.~\ref{lemmaind}, we will first show how its results can be used to prove Thm.~\ref{theoremproof}:

\begin{proof}[\emph{\textbf{Proof of Theorem \ref{theoremproof}}}]
As shown in  Appx.~\ref{sec:appxfN}, we can write
\begin{equation} \label{eq:appxfinal}
q(f_n^L) = \int \left[ \int q(f_M)\prod_{l=1}^L p(f_n^l | f_M^l; f_n^{l-1}) df_M \right] df_n^{1}\cdots df_n^{L-1}.
\end{equation}
Obtaining a formula for the inner integral can be done using Lem.~\ref{lemmaind} with $ l=L $, in which case Eq.~\eqref{eq:indmain} reads
\begin{equation}
\int q(f_M^1, \dots, f_M^L)\prod_{l'=1}^L p(f_n^{l'} | f_M^{l'}) df_M^{l'} =
\left[ \prod_{l'=1}^{L-1} q( f_n^{l'} | f_n^1, \dots, f_n^{l'-1})\right]
q(f_n^L | f_n^1,\dots, f_n^{L-1})
\end{equation}
since there is nothing left to integrate over. According to Eqs.~\eqref{eq:dgpsamplingfc} and \eqref{eq:indgauss} the distribution $ q(f_n^L | f_n^1,\dots, f_n^{L-1}) $ has the form necessary to be written as part of the product and we therefore have
\begin{equation}
\int q(f_M^1, \dots, f_M^L)\prod_{l'=1}^L p(f_n^{l'} | f_M^{l'}) df_M^{l'} =
\prod_{l'=1}^{L} q( f_n^{l'} | f_n^1, \dots, f_n^{l'-1}).
\end{equation}
Plugging this into Eq.~\eqref{eq:appxfinal} yields
\begin{equation} \label{eq:appxfinal2}
q(f_n^L) = \int \prod_{l'=1}^{L} q( f_n^{l'} | f_n^1, \dots, f_n^{l'-1}) df_n^{1}\cdots df_n^{L-1},
\end{equation}
where the distributions $ q $ on the right hand side have the properties described in Eqs.~\eqref{eq:dgpsamplingfc} - \eqref{eq:Sigmahatn}.
\end{proof}

In order to prove Lem.~\ref{lemmaind}, we will regularly need two standard formulas from Gaussian calculus, namely conditioning multivariate Gaussians,
\begin{equation}
\label{eq:gacond}
\gauss{\begin{pmatrix}
	x\\ y
	\end{pmatrix}}{\begin{pmatrix}
	a \\ b
	\end{pmatrix}}{\begin{pmatrix}
	A & C \\
	C^\top & B
	\end{pmatrix}} = \gauss{x}{a}{A}\gauss{y}{b + C^\top A^{-1}(x-a)}{B - C^\top A^{-1} C},
\end{equation}
and solving Gaussian integrals ("propagation"):
\begin{equation}
\label{eq:gaint}
\int\gauss{x}{a + Fy}{A}\gauss{y}{b}{B} dy = \gauss{x}{a + Fb}{A +FBF^\top}.
\end{equation}

\begin{proof}[\emph{\textbf{Proof of Lemma \ref{lemmaind}}}]
	As already said, we prove the lemma by induction:
	\paragraph{Base case}
	We need to show that Eq.~\eqref{eq:indmain} holds for $ l=1 $, i.e., that
	\begin{equation}\label{eq:indbase}
	\int q(f_M^1, \dots, f_M^L)\prod_{l'=1}^L p(f_n^{l'} | f_M^{l'}) df_M^{l'} =
	\int q(f_n^1,f_M^2,\dots,f_M^L ) \prod_{l'=2}^L p(f_n^{l'} | f_M^{l'}) df_M^{l'},
	\end{equation}
	where $ q(f_n^1,f_M^2,\dots,f_M^L ) $ is given according to Eqs.~\eqref{eq:indgauss} and \eqref{eq:indhelp}.
	
	In order to do so, we will perform the following steps:
	\begin{enumerate}[i)]
		\item Starting with the LHS of Eq.~\eqref{eq:indbase}, we isolate all terms that depend on $ f_M^1 $:
		\begin{equation}
		\int\left[ \int q(f_M^1, \dots, f_M^L) p(f_n^{1} | f_M^{1}) df_M^{1}\right]\prod_{l'=2}^L p(f_n^{l'} | f_M^{l'}) df_M^{l'}.
		\end{equation}
		\item In the previous equation, we only consider the inner integral and condition $ q $ on $ f^1_M $:
		\begin{equation}
		\int q(f_M^1, \dots, f_M^L) p(f_n^{1} | f_M^{1}) df_M^{1} = \int q(f_M^1) q(f_M^2, \dots, f_M^L|f_M^1) p(f_n^{1} | f_M^{1}) df_M^{1}.
		\end{equation}
		\item Next, we obtain the joint distribution of the two terms that are conditioned on $ f_M^1 $:
		\begin{equation}
		\int q(f_M^1) q(f_M^2, \dots, f_M^L|f_M^1) p(f_n^{1} | f_M^{1}) df_M^{1} = \int q(f_M^1) q(f_n^{1},f_M^2, \dots, f_M^L|f_M^1)  df_M^{1}.
		\end{equation}
		\item Then we evaluate the integral:
		\begin{equation}
		\int q(f_M^1) q(f_n^{1},f_M^2, \dots, f_M^L|f_M^1)  df_M^{1} = q(f_n^{1},f_M^2, \dots, f_M^L).
		\end{equation}
		\item Finally, we check that the resulting distribution is given by Eqs.~\eqref{eq:indgauss} and \eqref{eq:indhelp}. This then proves the equality in Eq.~\eqref{eq:indbase}.
	\end{enumerate}

	Step ii) is the first one where we actually need to calculate something, namely the conditioning of $ q(f_M^1, \dots, f_M^L) $. Using its definition in Eq.~\eqref{eq:appxqfM} and performing the conditioning according to Eq.~\eqref{eq:gacond} yields
	\begin{equation}
	\label{eq:baseii}
	\begin{aligned}
	&q(f_M^1, \dots, f_M^L) = q(f_M^1) q(f_M^2, \dots, f_M^L | f_M^1) \\
	=\ &\gauss{f_M^1}{\mu_M^1}{S_M^{11}}
	\gauss{f_M^{2:L}}{\mu_M^{2:L} + \SM{2:L}{1} \left(S_M^{11}\right)^{-1} (f_M^1-\mu_M^1) }{\SM{2:L}{2:L} - \SM{2:L}{1} \left(S_M^{11}\right)^{-1} \SM{1}{2:L} }.
	\end{aligned}
	\end{equation}
	
	For step iii) we use the formula we just obtained for $ q(f_M^2, \dots, f_M^L | f_M^1) $ and additionally $ p(f_n^{1} | f_M^{1}) $, which, according to Eq.~\eqref{eq:appxpcond} is given by $ \gauss{f_n^{1}}{\widetilde{\mathcal{K}}_{nM}^1 f_M^1 }{ \widetilde{ \mathcal{K} }_{nn}^1} $, and then proceed to build their joint Gaussian distribution:
	\begin{equation}\label{eq:baseiii}
	\begin{aligned}
	&q(f_n^1,f_M^2, \dots, f_M^L | f_M^1) = 
	p(f_n^{1} | f_M^{1}) q(f_M^2, \dots, f_M^L | f_M^1) \\
	=\ &\gauss{\begin{pmatrix*}[l]
		f_n^{1} \\ f_M^{2:L}
		\end{pmatrix*}}{\begin{pmatrix}
		\widetilde{\mathcal{K}}_{nM}^1 f_M^1 \\ \mu_M^{2:L} + \SM{2:L}{1} \left(S_M^{11}\right)^{-1} (f_M^1-\mu_M^1)
		\end{pmatrix}}{\begin{pmatrix}
		\widetilde{\mathcal{K}}_{nn}^1 & 0 \\
		0 & \SM{2:L}{2:L} - \SM{2:L}{1} \left(S_M^{11}\right)^{-1} \SM{1}{2:L}
		\end{pmatrix}}.
	\end{aligned}
	\end{equation}
	
	In step iv) we perform the integration using the term above for the joint and $ q(f_M^1) = \gauss{f_M^1}{\mu_M^1}{S_M^{11}} $ from Eq.~\eqref{eq:baseii} for the marginal. Applying Eq.~\eqref{eq:gaint} yields
	\begin{equation}\label{eq:baseiv}
	\begin{aligned}
	&\int q(f_n^1,f_M^2,\dots,f_M^L | f_M^{1}) q(f_M^1) df_M^{1} \\
	=\ &\mathcal{N}
	\left(
	\begin{pmatrix*}[l]	f_n^{1} \\ f_M^{2:L}\end{pmatrix*}
	\middle| \vphantom{\begin{pmatrix}
		\widetilde{\mathcal{K}}_{nM}^1 \mu_M^1 \\ \mu_M^{2:L} + \SM{2:L}{1} \left(S_M^{11}\right)^{-1} (\mu_M^1-\mu_M^1)
		\end{pmatrix}} \right.
	\begin{aligned}[t]
	&\begin{pmatrix}
	\widetilde{\mathcal{K}}_{nM}^1 \mu_M^1 \\ \mu_M^{2:L} + \SM{2:L}{1} \left(S_M^{11}\right)^{-1} (\mu_M^1-\mu_M^1)
	\end{pmatrix}, \\
	&\left. \begin{pmatrix}
	\widetilde{\mathcal{K}}_{nn}^1 & 0 \\
	0 & \SM{2:L}{2:L} - \SM{2:L}{1} \left(S_M^{11}\right)^{-1} \SM{1}{2:L}
	\end{pmatrix} +
	\begin{pmatrix}
	\widetilde{\mathcal{K}}_{nM}^1 \\ \SM{2:L}{1} \left(S_M^{11}\right)^{-1}
	\end{pmatrix} S_M^{11} \begin{pmatrix}
	\widetilde{\mathcal{K}}_{nM}^1 \\ \SM{2:L}{1} \left(S_M^{11}\right)^{-1}
	\end{pmatrix}^\top \right)
	\end{aligned}\\
	=\ &\gauss{\begin{pmatrix*}[l]	f_n^{1} \\ f_M^{2:L}\end{pmatrix*}}{\begin{pmatrix}
		\widetilde{\mu}_n^1 \\[3pt] \mu_M^{2:L}
		\end{pmatrix}}{\begin{pmatrix*}
		\widetilde{S}_n^{11} & \widetilde{\mathcal{K}}_{nM}^1\SM{1}{2:L} \\[3pt]
		\SM{2:L}{1}\widetilde{\mathcal{K}}_{Mn}^1 & \SM{2:L}{2:L}
		\end{pmatrix*}}.
	\end{aligned}
	\end{equation}
	In order to arrive at the last line we simplified the terms and used the definitions of $ \widetilde{\mu}_n^1 $ and $ \widetilde{S}_n^{11} $ in Thm.~\ref{theoremproof}.
	
	Step v) requires us to evaluate Eq.~\eqref{eq:indgauss} for $ l=1 $ resulting in
	\begin{equation}
	\gauss{\begin{pmatrix*}[l]
		f_n^1 \\ f_M^{2:L}
		\end{pmatrix*}}{\begin{pmatrix*}[l]
		\phantom{^1}\hat{\mu}_n^1 \\ ^1\hat{\mu}_M^{2:L}
		\end{pmatrix*}}{\begin{pmatrix*}
		\hat{\Sigma}^1_n & ^1\hat{\Sigma}_{nM}^{1,2:L} \\
		\left(^1\hat{\Sigma}_{nM}^{l,2:L}\right)^\top & ^1\hat{\Sigma}_{M}^{2:L,2:L}
		\end{pmatrix*}},
	\end{equation}
	which is the term $ q(f_n^1,f_M^2,\dots,f_M^L) $ on the RHS of Eq.~\eqref{eq:indbase}.
	Plugging in the definitions from Eq.~\eqref{eq:indhelp} we can easily verify that this last term indeed agrees with Eq.~\eqref{eq:baseiv}.
	Therefore our statement in Lem.~\ref{lemmaind} holds for $ l=1 $.
	
	\paragraph{Inductive step}
	We assume that Lemma \ref{lemmaind} holds for some $ l=1,\dots,L-1 $ (induction hypothesis) and then need to show that it also holds for $ l+1 $.
	That is, assuming that 
	\begin{equation}\label{eq:steprepeat}
	\int q(f_M^{1:L})\prod_{l'=1}^L p(f_n^{l'} | f_M^{l'}) df_M^{l'} =
	\left[ \prod_{l'=1}^{l-1} q( f_n^{l'} | f_n^{1:l'-1})\right]
	\int q(f_n^l,f_M^{l+1:L} | f_n^{1:l-1}) \prod_{l'=l+1}^L p(f_n^{l'} | f_M^{l'}) df_M^{l'},
	\end{equation}
	holds for some $ l $ with the terms on the RHS given by Eqs.~\eqref{eq:indgauss}, and \eqref{eq:indhelp} we need to show that we can also write the previous equation as
	\begin{equation} \label{eq:stepmain}
	\left[ \prod_{l'=1}^{l} q( f_n^{l'} | f_n^{1:l'-1})\right]
	\int q(f_n^{l+1},f_M^{l+2:L} | f_n^{1:l}) \prod_{l'=l+2}^L p(f_n^{l'} | f_M^{l'}) df_M^{l'},
	\end{equation}
	where this time the terms are given by Eqs.~\eqref{eq:indgauss}, and \eqref{eq:indhelp} but with $ l\to l+1 $.
	
	The way to show this is very similar to the way we showed the base case, the resulting formulas will only look more complicated and we will need one additional step in the beginning:
	\begin{enumerate}[i)]
		\item[o)] Assuming that Eq.~\eqref{eq:steprepeat} holds for some $ l $, we can start immediately with the RHS.
		The first step will be to marginalise $ f_n^l $ from the distribution $ q $ within the integral and show that the resulting marginal $ q(f_n^l| f_n^{1:l-1}) $ has the right form to be written as part of the product in front of the integral:
		\begin{align} \label{eq:stepo1}
			&\left[ \prod_{l'=1}^{l-1} q( f_n^{l'} | f_n^{1:l'-1})\right]
			\int q(f_n^l,f_M^{l+1:L} | f_n^{1:l-1}) \prod_{l'=l+1}^L p(f_n^{l'} | f_M^{l'}) df_M^{l'} \\ \label{eq:stepo2}
			=\ &\left[ \prod_{l'=1}^{l-1} q( f_n^{l'} | f_n^{1:l'-1})\right]
			\int q(f_n^l | f_n^{1:l-1}) q(f_M^{l+1:L} | f_n^{1:l}) \prod_{l'=l+1}^L p(f_n^{l'} | f_M^{l'}) df_M^{l'} \\ \label{eq:stepo3}
			=\ &\left[ \prod_{l'=1}^{l} q( f_n^{l'} | f_n^{1:l'-1})\right]
			\int q(f_M^{l+1:L} | f_n^{1:l}) \prod_{l'=l+1}^L p(f_n^{l'} | f_M^{l'}) df_M^{l'}.
		\end{align}
		Having done this, we will have to do the exact same steps as in the base case, which we will repeat below with updated indices.
	\end{enumerate}
	\begin{enumerate}[i)]
		\item Continuing from Eq.~\eqref{eq:stepo3}, we isolate all terms that depend on $ f_M^{l+1} $:
		\begin{equation}
		\left[ \prod_{l'=1}^{l} q( f_n^{l'} | f_n^{1:l'-1})\right]
		\int \left[ \int  q(f_M^{l+1:L} | f_n^{1:l}) p(f_n^{l+1} | f_M^{l+1}) df_M^{l+1} \right] \prod_{l'=l+2}^L p(f_n^{l'} | f_M^{l'}) df_M^{l'}.
		\end{equation}
		\item Comparing this to Eq.~\eqref{eq:stepmain}, we see that it remains to be shown that the inner integral equals $  q(f_n^{l+1},f_M^{l+2:L} | f_n^{1:l}) $ [given by Eqs.~\eqref{eq:indgauss} and \eqref{eq:indhelp}].
		Therefore we only consider the inner integral and therein condition $ q $ on $ f_M^{l+1} $:
		\begin{equation}
		\int  q(f_M^{l+1:L} | f_n^{1:l}) p(f_n^{l+1} | f_M^{l+1}) df_M^{l+1} = \int  q(f_M^{l+1}| f_n^{1:l}) q(f_M^{l+2:L} | f_n^{1:l},f_M^{l+1}) p(f_n^{l+1} | f_M^{l+1}) df_M^{l+1}.
		\end{equation}
		\item Next, we obtain the joint distribution of the two terms that are conditioned on $ f_M^{l+1} $:
		\begin{equation}
		\int  q(f_M^{l+1}| f_n^{1:l}) q(f_M^{l+2:L} | f_n^{1:l},f_M^{l+1}) p(f_n^{l+1} | f_M^{l+1}) df_M^{l+1} = \int  q(f_M^{l+1}| f_n^{1:l}) q(f_n^{l+1} ,f_M^{l+2:L} | f_n^{1:l},f_M^{l+1}) df_M^{l+1}.
		\end{equation}
		\item Then we evaluate the integral:
		\begin{equation}
		\int  q(f_M^{l+1}| f_n^{1:l}) q(f_n^{l+1} ,f_M^{l+2:L} | f_n^{1:l},f_M^{l+1}) df_M^{l+1} = q(f_n^{l+1} ,f_M^{l+2:L} | f_n^{1:l}).
		\end{equation}
		\item Finally, we check that the resulting distribution is given by Eqs.~\eqref{eq:indgauss} and \eqref{eq:indhelp}.
		This then proves the equality of Eqs.~\eqref{eq:steprepeat} and \eqref{eq:stepmain}.
	\end{enumerate}

	Let us begin with step o): According to Eq.~\eqref{eq:indgauss}, we have
	\begin{equation}
	q(f_n^l,f_M^{l+1:L} | f_n^{1:l-1}) = \gauss{\begin{pmatrix*}[l]
		f_n^l \\ f_M^{l+1:L}
		\end{pmatrix*}}{\begin{pmatrix*}[l]
		\phantom{^l}\hat{\mu}_n^l \\ ^l\hat{\mu}_M^{l+1:L}
		\end{pmatrix*}}{\begin{pmatrix*}
		\hat{\Sigma}^l_n & ^l\hat{\Sigma}_{nM}^{l,l+1:L} \\
		\left(^l\hat{\Sigma}_{nM}^{l,l+1:L}\right)^\top & ^l\hat{\Sigma}_{M}^{l+1:L,l+1:L}
		\end{pmatrix*}},
	\end{equation}
	which we condition on $ f_n^l $ using Eq.~\eqref{eq:gacond} (i.e., going from Eq.~\eqref{eq:stepo1} to Eq.~\eqref{eq:stepo2}):
	\begin{equation}	
	\begin{aligned} \label{eq:stepoexplicit}
	&q(f_n^l,f_M^{l+1:L} | f_n^{1:l-1}) = q(f_n^l | f_n^{1:l-1}) q(f_M^{l+1:L} | f_n^{1:l})
	= \gauss{f_n^l}{\hat{\mu}_n^l}{\hat{\Sigma}_n^l} \times \\ &\gauss{f_M^{l+1:L}}{\muM{l+1:L} + \SigMn{l+1:L}{l} \Signinv (f_n^l - \hat{\mu}_n^l) }{\SigM{l+1:L}{l+1:L} - \SigMn{l+1:L}{l} \Signinv \SignM{l}{l+1:L}}.
	\end{aligned}
	\end{equation}
	We therefore see that $ q(f_n^l | f_n^{1:l-1}) = \gauss{f_n^l}{\hat{\mu}_n^l}{\hat{\Sigma}_n^l} $, which is the right form for it to be included in the product in front of the integral in Eq.~\eqref{eq:stepo2}.
	This lets us arrive at Eq.~\eqref{eq:stepo3}, hence finishing step o).
	
	In step i) nothing really happens, we just note that, according to Eq.~\eqref{eq:appxpcond},
	\begin{equation} \label{eq:stepi}
	p(f_n^{l+1} | f_M^{l+1}) = \gauss{f_n^{l+1}}{\widetilde{ \mathcal{K}}_{nM}^{l+1} f_M^l}{ \widetilde{ \mathcal{K}}_{nn}^{l+1}}.
	\end{equation}
	
	Using $ q(f_M^{l+1:L} | f_n^{1:l}) $ from Eq.~\eqref{eq:stepoexplicit}, we perform step ii) according to Eq.~\eqref{eq:gacond}, resulting in
	\begin{equation}
	q(f_M^{l+1:L} | f_n^{1:l}) = q(f_M^{l+1}| f_n^{1:l}) q(f_M^{l+2:L} | f_n^{1:l},f_M^{l+1}),
	\end{equation}
	where
	\begin{equation}\label{eq:stepii1}
	q(f_M^{l+1} | f_n^{1:l}) = \gauss{f_M^{l+1}}{\muM{l+1} + \SigMn{l+1}{l} \Signinv (f_n^l - \hat{\mu}_n^l) }{\SigM{l+1}{l+1} - \SigMn{l+1}{l} \Signinv \SignM{l}{l+1}}
	\end{equation}
	and
	\begin{equation}\label{eq:stepii2}
	\begin{aligned}
	&q(f_M^{l+2:L} | f_n^{1:l}, f_M^{l+1}) \\
	=\ &\begin{aligned}[t]
	\mathcal{N}\left( f_M^{l+2:L} \middle| \vphantom{\Signinv} \right. &\muM{l+2:L} + \SigMn{l+2:L}{l} \Signinv (f_n^l - \hat{\mu}_n^l) + \left( \SigM{l+2:L}{l+1} - \SigMn{l+2:L}{l} \Signinv \SignM{l}{l+1} \right) \times \\
	&\left( \SigM{l+1}{l+1} - \SigMn{l+1}{l} \Signinv \SignM{l}{l+1} \right)^{-1} \left( f_M^{l+1} - \muM{l+1} - \SigMn{l+1}{l} \Signinv (f_n^l - \hat{\mu}_n^l) \right),\\
	&\SigM{l+2:L}{l+2:L} - \SigMn{l+2:L}{l} \Signinv \SignM{l}{l+2:L} - \left( \SigM{l+2:L}{l+1} - \SigMn{l+2:L}{l} \Signinv \SignM{l}{l+1} \right) \times \\
	&\left( \SigM{l+1}{l+1} - \SigMn{l+1}{l} \Signinv \SignM{l}{l+1} \right)^{-1} \left( \SigM{l+1}{l+2:L} - \SigMn{l+1}{l} \Signinv \SignM{l}{l+2:L}\right) \left. \vphantom{\Signinv} \right).
	\end{aligned}
	\end{aligned}
	\end{equation}
	
	For step iii) we have to build the joint Gaussian distribution
	\begin{equation} \label{eq:stepiii}
	q(f_M^{l+2:L} | f_n^{1:l},f_M^{l+1}) p(f_n^{l+1} | f_M^{l+1}) = q(f_n^{l+1} ,f_M^{l+2:L} | f_n^{1:l},f_M^{l+1})
	\end{equation}
	using Eqs.~\eqref{eq:stepi} and \eqref{eq:stepii2}.
	Since this formula would be even longer than the one in Eq.~\eqref{eq:stepii2}, we refrain from explicitly writing it here.
	While the corresponding formula for the base case [Eq.~\eqref{eq:baseiii}] is much simpler the resulting form of Eq.~\eqref{eq:stepiii} would be similar.
	
	Next, the integration in step iv) can be performed using Eqs.~\eqref{eq:gaint}, \eqref{eq:stepii1}, and \eqref{eq:stepiii}.
	The calculations are again very similar to the ones in the corresponding step for the base case [Eq.~\eqref{eq:baseiv}] so we only state the final result here:
	\begin{equation}\label{eq:stepiv}
	\begin{aligned}
	&q(f_n^{l+1} ,f_M^{l+2:L} | f_n^{1:l}) = \int q(f_M^{l+1}| f_n^{1:l}) q(f_n^{l+1} ,f_M^{l+2:L} | f_n^{1:l},f_M^{l+1}) df_M^{l+1} \\
	=\ &\gauss{\begin{pmatrix*}[l]
		f_n^{l+1} \\[2mm] f_M^{l+2:L}
		\end{pmatrix*}}{\begin{pmatrix*}[l]
		\hat{m}_n^{l+1} \\[2mm] \hat{m}_M^{l+2:L}
		\end{pmatrix*}}{\begin{pmatrix*}
		\hat{S}^{l+1}_n & \hat{S}_{nM}^{l+1,l+2:L} \\
		\left(\hat{S}_{nM}^{l+1,l+2:L}\right)^\top & \hat{S}_{M}^{l+2:L,l+2:L}
		\end{pmatrix*}},
	\end{aligned}	
	\end{equation}
	where
	\begin{align}\label{eq:stepresfirst}
	\hat{m}_n^{l+1} &= \widetilde{\mathcal{K}}_{nM}^{l+1}\left(\muM{l+1} + \SigMn{l+1}{l} \Signinv (f_n^l - \hat{\mu}_n^l) \right)\\\label{eq:stepressecond}
	\hat{m}_M^{l+2:L} &= \muM{l+2:L} + \SigMn{l+2:L}{l} \Signinv (f_n^l - \hat{\mu}_n^l)\\
	\hat{S}^{l+1}_n &= \mathcal{K}^{l+1}_{nn} + \mathcal{K}^{l+1}_{nM}\left( \SigM{l+1}{l+1} - \SigMn{l+1}{l} \Signinv \SignM{l}{l+1} \right) \mathcal{K}^{l+1}_{Mn}\\
	\hat{S}_{nM}^{l+1,l+2:L} &= \widetilde{\mathcal{K}}_{nM}^{l+1} \left( \SigM{l+1}{l+2:L} - \SigMn{l+1}{l} \Signinv \SignM{l}{l+2:L}\right)\\ \label{eq:stepreslast}
	\hat{S}_{M}^{l+2:L,l+2:L} &= \SigM{l+2:L}{l+2:L} - \SigMn{l+2:L}{l} \Signinv \SignM{l}{l+2:L}.
	\end{align}
	
	What remains to be shown in step v) is that this result does in fact agree with the expected result from Lem.~\ref{lemmaind}, i.e.,
	\begin{equation}\label{eq:stepremainRHS}
	q(f_n^{l+1},f_M^{l+2:L} | f_n^{1:l}) = \gauss{\begin{pmatrix*}[l]
		f_n^{l+1} \\ f_M^{l+2:L}
		\end{pmatrix*}}{\begin{pmatrix*}[l]
		\phantom{^{l+1}}\hat{\mu}_n^{l+1} \\ ^{l+1}\hat{\mu}_M^{l+2:L}
		\end{pmatrix*}}{\begin{pmatrix*}
		\hat{\Sigma}^{l+1}_n & ^{l+1}\hat{\Sigma}_{nM}^{l+1,l+2:L} \\
		\left(^{l+1}\hat{\Sigma}_{nM}^{l+1,l+2:L}\right)^\top & ^{l+1}\hat{\Sigma}_{M}^{l+2:L,l+2:L}
		\end{pmatrix*}},
	\end{equation}
	where the terms are defined in Eqs.~\eqref{eq:muhatn}, \eqref{eq:Sigmahatn}, and \eqref{eq:indhelp}.
	That means we have to prove that $ \hat{m}_n^{l+1} = \hat{\mu}_n^{l+1} $ and similarly for the other terms in Eqs.~\eqref{eq:stepressecond} - \eqref{eq:stepreslast}.
	Note that this is the point where we need the left indices in order to distinguish e.g.~the term $ \muM{l+2:L} $ appearing in Eq.~\eqref{eq:stepressecond} from $ ^{l+1}\hat{\mu}_M^{l+2:L} $ appearing in the mean of Eq.~\eqref{eq:stepremainRHS}.
	
	We will exemplarily prove that $ \hat{m}_n^{l+1} = \hat{\mu}_n^{l+1} $: Starting from Eq.~\eqref{eq:stepresfirst} we have 
	\begin{equation}\label{eq:stepmean1}
	\begin{aligned}
	&\hat{m}_n^{l+1} = \widetilde{\mathcal{K}}_{nM}^{l+1}\left(\muM{l+1} + \SigMn{l+1}{l} \Signinv (f_n^l - \hat{\mu}_n^l) \right)\\
	=\ &\widetilde{\mu}_n^{l+1} + \Sn{l+1}{1:l-1} \Sninv \left(f_n^{1:l-1} - \widetilde{\mu}_n^{1:l-1}\right) + \left( \Sn{l+1}{l} - \Sn{l+1}{1:l-1} \Sninv \Sn{1:l-1}{l} \right) \times \\ 
	&\Signinv \left( f_n^l - \widetilde{\mu}_n^{l}  - \Sn{l}{1:l-1} \Sninv \left(f_n^{1:l-1} - \widetilde{\mu}_n^{1:l-1}\right) \right),
	\end{aligned}
	\end{equation}
	where we used the definitions in Eqs.~\eqref{eq:muhatn} and \eqref{eq:indhelp} for the terms $ \hat{\cdot} $.
	Note that these definitions are part of the induction hypothesis.
	It will soon become clear why we did not substitute $ \hat{\Sigma}_n^l $.
	We furthermore used the definitions of the $ \widetilde{\mu}_n $ and $ \widetilde{S}_n $ terms in Thm.~\ref{theoremproof} to absorb the $ \widetilde{\mathcal{K}} $ terms.
	In the following we are going to write Eq.~\eqref{eq:stepmean1} in a vectorized form and additionally substitute
	\begin{equation}\label{eq:stepmeandef1}
	A = \Sn{1:l-1}{1:l-1} , \qquad  B = \Sn{1:l-1}{l}, \qquad C = \Sn{l}{1:l-1},
	\qquad \widetilde{D} =\hat{\Sigma}^l_n,
	\end{equation}
	 The reason for these steps will become clear after two more equations:
	\begin{equation}\label{eq:stepmean2}
	\hat{m}_n^{l+1} = \widetilde{\mu}_n^{l+1} +
	\begin{pmatrix*}[c]
	\Sn{l+1}{1:l-1} A^{-1} - \left( \Sn{l+1}{l} - \Sn{l+1}{1:l-1} A^{-1} B \right) \widetilde{D}^{-1} C A^{-1} \\
	\left( \Sn{l+1}{l} - \Sn{l+1}{1:l-1} A^{-1} B \right) \widetilde{D}^{-1}
	\end{pmatrix*}^\top
	\begin{pmatrix}
	f_n^{1:l-1} - \widetilde{\mu}_n^{1:l-1} \\ f_n^l - \widetilde{\mu}_n^l
	\end{pmatrix}
	\end{equation}
	Going one step further, we recognize this as a vector matrix multiplication,
	\begin{equation}\label{eq:stepmean3}	
	\hat{m}_n^{l+1} = \widetilde{\mu}_n^{l+1} + \begin{pmatrix}
	\Sn{l+1}{1:l-1} \\ \Sn{l+1}{l}
	\end{pmatrix}^\top 
	\begin{pmatrix}
	A^{-1} + A^{-1}B\widetilde{D}^{-1}CA^{-1} & -A^{-1}B\widetilde{D}^{-1} \\
	-\widetilde{D}^{-1}CA^{-1} & \widetilde{D}^{-1}
	\end{pmatrix}
	\begin{pmatrix}
	f_n^{1:l-1} - \widetilde{\mu}_n^{1:l-1} \\ f_n^l - \widetilde{\mu}_n^l
	\end{pmatrix},
	\end{equation}
	where we additionally exploited that $ A $ and $ \widetilde{D} $ are symmetric and that $ B^\top = C $.
	In order to get any further from here we need the block matrix inversion lemma, which states that
	\begin{equation}
	\label{eq:blockinv}
	\begin{pmatrix}
	A & B \\
	C & D
	\end{pmatrix}^{-1} = 
	\begin{pmatrix}
	A^{-1} + A^{-1}B\widetilde{D}^{-1}CA^{-1} & -A^{-1}B\widetilde{D}^{-1} \\
	-\widetilde{D}^{-1}CA^{-1} & \widetilde{D}^{-1}
	\end{pmatrix},
	\end{equation}
	where $\widetilde{D} = D - CA^{-1}B$. Comparing Eqs.~\eqref{eq:stepmean3} and \eqref{eq:blockinv} explains why we insisted on vectorising the last few formulas and also our definitions in Eq.~\eqref{eq:stepmeandef1}.
	Finally, since $\hat{\Sigma}_n^l = \widetilde{S}_n^{ll} - \widetilde{S}_n^{l,1:l-1} \left( \widetilde{S}_n^{1:l-1,1:l-1} \right)^{-1} \widetilde{S}_n^{1:l-1,l}$ [Eq.~\eqref{eq:Sigmahatn}], we also identify $ \widetilde{S}_n^{ll}= D $.
	We can therefore rewrite Eq.~\eqref{eq:stepmean3} by reversing the block matrix inversion and resubstituting the terms in Eq.~\eqref{eq:stepmeandef1}:
	\begin{equation}\label{eq:stepmeanfinal}
	\begin{aligned}
	\hat{m}_n^{l+1} &= \widetilde{\mu}_n^{l+1} + \begin{pmatrix}
	\Sn{l+1}{1:l-1} \\[2mm] \Sn{l+1}{l}
	\end{pmatrix}^\top \begin{pmatrix}
	\Sn{1:l-1}{1:l-1} & \Sn{1:l-1}{l} \\[2mm]
	\Sn{l}{1:l-1} & \widetilde{S}_n^{ll}
	\end{pmatrix}^{-1}
	\begin{pmatrix}
	f_n^{1:l-1} - \widetilde{\mu}_n^{1:l-1} \\[2mm] f_n^l - \widetilde{\mu}_n^l
	\end{pmatrix} \\
	&= \widetilde{\mu}_n^{l+1} + \Sn{l+1}{1:l} \left( \Sn{1:l}{1:l}\right)^{-1}
	\left( f_n^{1:l} - \widetilde{\mu}_n^{1:l}\right).
	\end{aligned}
	\end{equation}
	In the last step we simply rewrote the vectors and the matrix according to the way we defined the submatrix notation.
	Comparing the final result to Eq.~\eqref{eq:muhatn}, we realize that this is indeed $ \hat{\mu}^{l+1}_n $, i.e., the mean term where we substituted $ l\to l+1 $.
	In exactly the same way, i.e., by reversing the matrix inversion, we can show that the other parameters of the distribution in Eq.~\eqref{eq:stepiv} indeed coincide with the respective parameters of the distribution in Eq.~\eqref{eq:stepremainRHS}.
	Since this was the last part that remained to be shown, we finished the proof of Lem.~\ref{lemmaind}.
\end{proof}

%% file: subfiles/appendix_theory_intuition.tex
{\new{
\section{Intuition for the proof of Theorem 1}
In this section, we provide some intuition that might be helpful in understanding parts of the proof of Thm.~\ref{theoremproof}.
In the first part, we present a different, heuristic way of obtaining the same results, making use of a mathematically wrong (or at least not  mathematically rigorous) step.
This helped us come up with the exact form of the theorem that we proved above.
The second part gives some intuition on how the formulas appearing in Thm.~\ref{theoremproof}, especially Eqs.~\eqref{eq:muhatn} and \eqref{eq:Sigmahatn}, can be interpreted. More precisely, we show how these reduce to the mean-field equations when we plug in the mean-field covariance matrix.

\subsection{Heuristic argument for Theorem 1}
\label{sec:appxintuition1}
The aim of this section is to provide a heuristic argument for Thm.~\ref{theoremproof} as opposed to the complete proof given in Appx.~\ref{sec:appxproof}.
For convenience we recap the starting point of that section:
We need to show that
\begin{equation}\label{eq:appxheurmain}
\int q(f_M)\prod_{l=1}^L p(f_n^l | f_M^l; f_n^{l-1}) df_M = \prod_{l=1}^{L} q(f_n^l | f_n^1,\dots,f_n^{l-1}),
\end{equation}
where the distributions $ q $ on the right hand side are defined in Eqs.~\eqref{eq:dgpsamplingfc} - \eqref{eq:Sigmahatn}.
The terms appearing on the left hand side are given by $ q(f_M) = \gauss{f_M}{\mu_M}{S_M} $, which is interchangeably also denoted as
\begin{equation}\label{eq:appxheurqfM}
q(f_M^{1:L}) = \gauss{f_M^{1:L}}{\mu_M^{1:L}}{S_M^{1:L,1:L}} = q
\begin{pmatrix}
f_M^1\\
\vdots\\
f_M^L
\end{pmatrix} = \mathcal{N}
\left(
\begin{pmatrix}
f_M^1\\
\vdots\\
f_M^L
\end{pmatrix}
\middle|
\begin{pmatrix}
\mu_M^1\\
\vdots\\
\mu_M^L
\end{pmatrix},
\begin{pmatrix}
S_M^{11} & \cdots & S_M^{1L}\\
\vdots & \ddots & \vdots\\
S_M^{L1} & \cdots & S_M^{LL}
\end{pmatrix}
\right),
\end{equation}
and
\begin{equation}\label{eq:appxpheurcond}
p(f_n^l | f_M^l; f_n^{l-1}) = \gauss{ f_n^l }{ \widetilde{\mathcal{K}}_{nM}^l f_M^l}{ \widetilde{\mathcal{K}}_{nn}^l },
\end{equation}
where
\begin{alignat}{2}\label{eq:appxheurKnM}
&\widetilde{\mathcal{K}}_{nM}^l &&= \mathcal{K}_{nM}^l \left( \mathcal{K}_{MM}^l \right)^{-1} \\ \label{eq:appxheurKnn}
&\widetilde{\mathcal{K}}_{nn}^l &&= \mathcal{K}_{nn}^l - \mathcal{K}_{nM}^l \left( \mathcal{K}_{MM}^l \right)^{-1} \mathcal{K}_{Mn}^l.
\end{alignat}

Let us consider $\prod_{l=1}^L p(f_n^l | f_M^l; f_n^{l-1})$ on the left hand side of Eq.~\eqref{eq:appxheurmain} as a joint multivariate distribution,
\begin{equation}\label{eq:appxheurpfn}
p(f_n | f_M) =
\mathcal{N}
\left(
\begin{pmatrix}
f_n^1\\
\vdots\\
f_n^L
\end{pmatrix}
\middle|
\begin{pmatrix}
\widetilde{\mathcal{K}}_{nM}^1 & 0 & 0\\
0 & \ddots & 0\\
0 & 0 & \widetilde{\mathcal{K}}_{nM}^L
\end{pmatrix}
\begin{pmatrix}
f_M^1\\
\vdots\\
f_M^L
\end{pmatrix},
\begin{pmatrix}
\widetilde{\mathcal{K}}_{nn}^1 & 0 & 0\\
0 & \ddots & 0\\
0 & 0 & \widetilde{\mathcal{K}}_{nn}^L
\end{pmatrix}
\right).
\end{equation}
Note that this is the point where this ``proof'' becomes heuristic:
The object on the right hand side of Eq.~\eqref{eq:appxheurpfn} is not really a probability distribution, as the variables over which the distribution is defined (the $ f_n^l $) appear as parameters of the distribution itself (as inputs of the covariance matrices, e.g.~$ \widetilde{\mathcal{K}}_{nn}^{l+1} $ ).
In the following we will pretend that rules for (multivariate Gaussian) distributions still apply to this object, making the rest of this proof mathematically wrong.
We hope that it can still provide some intuition.
Using the standard formula for solving Gaussian integrals ("propagation"),
\begin{equation}
\label{eq:gaintheur}
\int\gauss{x}{a + Fy}{A}\gauss{y}{b}{B} dy = \gauss{x}{a + Fb}{A +FBF^\top},
\end{equation}
we can then easily plug in Eq.~\eqref{eq:appxheurpfn} in the left hand side of Eq.~\eqref{eq:appxheurmain}, yielding
\begin{equation}
\label{eq:appxheurhacky}
q(f_n) = \int q(f_M) p(f_n | f_M) df_M =
\gauss{
	\begin{pmatrix}
	f_n^1\\
	\vdots\\
	f_n^L
	\end{pmatrix}}
{
	\begin{pmatrix}
	\widetilde{\mu}_n^1\\
	\vdots\\
	\widetilde{\mu}_n^L
	\end{pmatrix}}
{
	\begin{pmatrix}
	\widetilde{S}_n^{11} & \cdots & \widetilde{S}_n^{1L}\\
	\vdots & \ddots & \vdots\\
	\widetilde{S}_n^{L1} & \cdots & \widetilde{S}_n^{LL}
	\end{pmatrix}},
\end{equation}
where
\begin{align}
\label{eq:appxheurmutilde}
\widetilde{\mu}_n^l &= \widetilde{\mathcal{K}}^l_{nM} \mu_M^l, 
\\
\label{eq:appxheurStilde}
\widetilde{S}_n^{ll'} &= \delta_{ll'} \mathcal{K}^l_{nn} - \widetilde{\mathcal{K}}^l_{nM} \left(\delta_{ll'} \mathcal{K}^l_{MM} - S_M^{ll'}\right) \widetilde{\mathcal{K}}^{l'}_{Mn}.
\end{align}
Note that the latter two definitions also appear in Thm.~\ref{theoremproof}.
The expression in Eq.~\eqref{eq:appxheurhacky} has still the same problem as the expression in Eq.~\eqref{eq:appxheurpfn} in that it is not a valid distribution (since $ \widetilde{\mu}_n^l $ and $ \widetilde{S}_n^{ll'} $ depend on $ f_n^{l-1} $).

Pretending further, that rules for distributions still apply, we can
use the standard formula for conditioning multivariate Gaussians,
\begin{equation}
\label{eq:gacondheur}
\gauss{\begin{pmatrix}
	x\\ y
	\end{pmatrix}}{\begin{pmatrix}
	a \\ b
	\end{pmatrix}}{\begin{pmatrix}
	A & C \\
	C^\top & B
	\end{pmatrix}} = \gauss{x}{a}{A}\gauss{y}{b + C^\top A^{-1}(x-a)}{B - C^\top A^{-1} C},
\end{equation}
to repeatedly condition Eq.~\eqref{eq:appxheurhacky}, resulting in
\begin{equation}\label{eq:heursamplingfc}
q(f_n) =  \prod_{l=1}^{L} q(f_n^l | f_n^1,\dots,f_n^{l-1}) 
\quad \text{where} \quad  q(f_n^l | f_n^1,\dots,f_n^{l-1}) = \gauss{f_n^l}{\hat{\mu}_n^l}{\hat{\Sigma}_n^l},
\end{equation}
and
\begin{align}
\label{eq:heurmuhatn}
\hat{\mu}_n^l \! &= \! \widetilde{\mu}_n^l + \widetilde{S}_n^{l,1:l-1} \left( \widetilde{S}_n^{1:l-1,1:l-1} \right)^{-1} ( f_n^{1:l-1} - \widetilde{\mu}_n^{1:l-1}), 
\\
\label{eq:heurSigmahatn}
\hat{\Sigma}_n^l \!  &= \! \widetilde{S}_n^{ll} - \widetilde{S}_n^{l,1:l-1} \left( \widetilde{S}_n^{1:l-1,1:l-1} \right)^{-1} \widetilde{S}_n^{1:l-1,l}. 
\end{align}
In Eqs.~\eqref{eq:heurmuhatn} and \eqref{eq:heurSigmahatn} the notation $ A^{l,1:l'} $ is used to index a submatrix of the variable $ A $, e.g.~${A}^{l,1:l'} = \left(A^{l,1} \cdots A^{l,l'} \right)$.

The final result, Eqs.~\eqref{eq:heursamplingfc} - \eqref{eq:heurSigmahatn}, is once again a valid distribution and exactly matches the outcome of the mathematically rigorous proof of Thm.~\ref{theoremproof} in Appx.~\ref{sec:appxproof}. The latter, while being much more complicated, is necessary since the intermediate expressions [Eqs.~\eqref{eq:appxheurpfn} and \eqref{eq:appxheurhacky}] rely on mathematically wrong (or at least dubious) steps.

\subsection{Mean-field as a structured approximation}
\label{sec:appxintuition2}
Here, we verify that when we plug in the mean-field covariance matrix into the formulas appearing in Thm.~\ref{theoremproof}, we recover the mean-field formulas appearing in Sec.~\ref{sec:mf}, i.e., that in this case Eq.~\eqref{eq:dgpsamplingfc} reduces to Eq.~\eqref{eq:dgpsamplingmf}.
For convenience we repeat the relevant formulas, starting with the mean-field formula for the marginals of the last layer [Eq.~\eqref{eq:dgpsamplingmf}]:
\begin{equation}\label{eq:appxmf}
q(f_n^L) = \int \prod_{l=1}^{L} q(f_n^l;f_n^{l-1})df_n^{1}\cdots df_n^{L-1},
\quad \text{where} \quad
q(f_n^l;f_n^{l-1}) = \prod_{t=1}^{T_l} \gauss{f_n^{l,t}}{\widetilde{\mu}_n^{l,t}}{\widetilde{\Sigma}_n^{l,t}},
\end{equation}
where the means and covariances of the Gaussians in this equation are given by:
\begin{equation}\label{eq:appxmfmeanvar}
{\widetilde{\mu}_n^{l,t}}   =  \widetilde{K}^l_{nM}\mu^{l,t}_M,
\quad
{\widetilde{\Sigma}_n^{l,t}} =  K^l_{nn} -
\widetilde{K}^l_{nM}\left(K^l_{MM} - S^{l,t}_M\right) \widetilde{K}^l_{Mn}.
\end{equation}
The formulas for the fully-coupled variant, starting with the marginals of the last layer [Eq.~\eqref{eq:dgpsamplingfc}], read,
\begin{equation}\label{eq:appxfc}
q(f_n^L) = \int \prod_{l=1}^{L} q(f_n^l | f_n^1,\dots,f_n^{l-1})  df_n^{1}\cdots df_n^{L-1}
\quad \text{where} \quad
q(f_n^l | f_n^1,\dots,f_n^{l-1}) = \gauss{f_n^l}{\hat{\mu}_n^l}{\hat{\Sigma}_n^l},
\end{equation}
where the means and covariances of the Gaussians in this equation are given by:
\begin{align}
\label{eq:appxfcmean}
\hat{\mu}_n^l \! &= \! \widetilde{\mu}_n^l + \widetilde{S}_n^{l,1:l-1} \left( \widetilde{S}_n^{1:l-1,1:l-1} \right)^{-1} ( f_n^{1:l-1} - \widetilde{\mu}_n^{1:l-1}), 
\\
\label{eq:appxfcvar}
\hat{\Sigma}_n^l \!  &= \! \widetilde{S}_n^{ll} - \widetilde{S}_n^{l,1:l-1} \left( \widetilde{S}_n^{1:l-1,1:l-1} \right)^{-1} \widetilde{S}_n^{1:l-1,l}, 
\end{align}
where
\begin{align}
\label{eq:appxfcmutilde}
\widetilde{\mu}_n^l &= \widetilde{\mathcal{K}}^l_{nM} \mu_M^l, 
\\
\label{eq:appxfcStilde}
\widetilde{S}_n^{ll'} &= \delta_{ll'} \mathcal{K}^l_{nn} - \widetilde{\mathcal{K}}^l_{nM} \left(\delta_{ll'} \mathcal{K}^l_{MM} - S_M^{ll'}\right) \widetilde{\mathcal{K}}^{l'}_{Mn}.
\end{align}
Here we introduced $ \mathcal{K}^l  = \left(\mathbb{I}_{T_l} \otimes K^l\right) $ as shorthand for the Kronecker product between the identity matrix $\mathbb{I}_{T_l}$ and the covariance matrix $ K^l $, and used $\delta$ for the Kronecker delta.

Having all relevant formulas in one place, we can proceed to show that if we plug in the mean field covariance matrix, $ S_M = \text{diag}(\{S_M^{ll}\}_{l=1}^L) $, where $ S_M^{ll} = \text{diag}(\{S_M^{l,t}\}_{t=1}^{T_l}) $ (see also Fig.~\ref{fig:covariance}, left), in Eq.~\eqref{eq:appxfc}, we recover Eq.~\eqref{eq:appxmf}:  
Removing the correlations between the layers by setting $ S_M = \text{diag}(\{S_M^{ll}\}_{l=1}^L) $,
also implies that $\tilde{S}_n^{ll'} = 0$ if $ l\neq l' $ [Eq.~\eqref{eq:appxfcStilde}]. Therefore Eqs.~\eqref{eq:appxfcmean} and \eqref{eq:appxfcvar} reduce to $ \hat{\mu}_n^l = \widetilde{\mu}_n^l $ and 
$\hat{\Sigma}_n^l = \widetilde{S}_n^{ll}$, respectively.
The resulting variational posterior factorises between the layers with $q(f_n^l ;f_n^{l-1}) = \gauss{f_n^l}{\widetilde{\mu}_n^l}{\widetilde{S}_n^{ll}}$.
Comparing with Eq.~\eqref{eq:appxmf}, we can already see that the means are equal [since $ \widetilde{\mu}_n^l = (\widetilde{\mu}_n^{l,1},\dots,\widetilde{\mu}_n^{l,T_l}) $, cf.~Eqs.~\eqref{eq:appxmfmeanvar} and \eqref{eq:appxfcmutilde}].
Removing the correlations within one layer by setting 
 $ S_M^{ll} = \text{diag}(\{S_M^{l,t}\}_{t=1}^{T_l}) $ renders the covariance matrix ${\widetilde{S}_n^{ll}}$ block-diagonal.
The diagonal blocks are obtained by evaluating Eq.~\eqref{eq:appxfcStilde} with $ S_M^{ll} = \text{diag}(\{S_M^{l,t}\}_{t=1}^{T_l}) $.
It is easy to see that these diagonal blocks are equal to $\widetilde{\Sigma}_n^{l,t}$ in Eq.~\eqref{eq:appxmfmeanvar} and we fully recover the mean-field solution [Eq.~\eqref{eq:appxmf}].
}}

%% file: subfiles/appendix_theory_part2.tex
\section{ELBO} \label{sec:appxelbo}
Here we show how to derive the ELBO of the FC DGP, i.e., Eq.~\eqref{eq:elbodgpfc}, which is given by
\begin{equation} \label{eq:appxelbofinal}
\mathcal{L} = \sum_{n=1}^{N}\mathbb{E}_{q(f_n^L)} \left[\log p(y_n|f_n^L)\right] - \text{KL}[q(f_M)||\prod_{l=1}^{L}p(f_M^l)].
\end{equation}
While this is already done in the supplemental material of Ref.~\cite{salimbeni-deisenroth-doubly-stochastic-vi-deep-gp}, we will do the derivation once again since our notation is different.
For convenience we repeat the relevant formulas from the main text, i.e., the general formula for the ELBO [Eq.~\eqref{eq:elbogeneral}], the joint DGP prior [Eq.~\eqref{eq:dgpprior}], and the variational family for the DGP [Eq.~\eqref{eq:dgpvarfam}], which are given by
\begin{align} \label{eq:appxelbogen}
\mathcal{L} &= \int q(f_N,f_M)\log \frac{p(y_N,f_N,f_M)}{q(f_N,f_M)} df_N df_M,\\
p(y_N,f_N,f_M) &= p(y_N|f_N^L)\prod_{l=1}^{L}p(f_N^l|f_M^l;f_N^{l-1})p(f_M^l),\\
q(f_N,f_M) &= q(f_M)\prod_{l=1}^{L}p(f_N^l|f_M^l;f_N^{l-1}), 
\end{align}
respectively. By only using the general form of the distributions, and exploiting that we assumed iid noise, i.e., $p(y_N|f_N^L) = \prod_{n=1}^N p(y_n|f_n^L)$, we can get from Eq.~\eqref{eq:appxelbogen} to Eq.~\eqref{eq:appxelbofinal}:
\begin{align}
\mathcal{L} &= \int q(f_N,f_M)\log \frac{p(y_N,f_N,f_M)}{q(f_N,f_M)} df_N df_M
 = \int q(f_N,f_M) \log \frac{p(y_N|f_N^L) \prod_{l=1}^{L} p(f_M^l)}{q(f_M)} df_N df_M\\
 &= \int q(f_N,f_M) \log p(y_N|f_N^L) df_N df_M + 
 \int q(f_N,f_M) \log \frac{\prod_{l=1}^{L} p(f_M^l)}{q(f_M)} df_N df_M,\\
 &= \int q(f_N,f_M) \log \prod_{n=1}^N p(y_n|f_n^L) df_N df_M + 
 \int q(f_M) \log \frac{\prod_{l=1}^{L} p(f_M^l)}{q(f_M)} df_M,\\
 &= \sum_{n=1}^{N} \int q(f_n^L) \log p(y_n|f_n^L) df_n^L -\text{KL}[q(f_M)||\prod_{l=1}^{L}p(f_M^l)].
 \end{align}
In the last step we introduced $ q(f_n^L) $  as simply summarising all remaining terms in the first integral, hence,
\begin{equation}\label{eq:appxqfnL}
q(f_n^L) = \int q(f_N,f_M) df_M \prod_{n'\neq n} df_{n'}^L \prod_{l=1}^{L-1} df_N^l.
\end{equation}

\section{Marginalisation of (most) latent layer outputs} \label{sec:appxfN}
Here we show how to get from the general form of $ q(f_n^L) $ given in Eq.~\eqref{eq:appxqfnL} to the starting point of our induction proof [Eq.~\eqref{eq:proofgen}], where all the latent outputs $ f_{n'}^l $ are integrated out for all layers $ l $ and for all samples $ n' \neq n $:
\begin{equation} \label{eq:appxqfnLproof}
q(f_n^L) = \int \left[ \int q(f_M)\prod_{l=1}^L p(f_n^l | f_M^l; f_n^{l-1}) df_M \right] df_n^{1}\cdots df_n^{L-1}.
\end{equation}

While this is already shown in Remark 2 in Ref.~\cite{salimbeni-deisenroth-doubly-stochastic-vi-deep-gp} (note that the indices there are not correct), we will provide a bit more detail here and we can also nicely point out where the difference in the formulas for $ q(f_n^L) $ arises from.
For convenience the relevant formulas are repeated below:
\begin{equation} \label{eq:appxreminder}
q(f_N,f_M) = q(f_M)\prod_{l=1}^{L}p(f_N^l|f_M^l;f_N^{l-1}), \qquad
p(f_N^l|f_M^l;f_N^{l-1}) = \gauss{ f_N^l }{ \widetilde{\mathcal{K}}_{NM}^l f_M^l }{ \widetilde{\mathcal{K}}_{NN}^l},
\end{equation}
where $\widetilde{\mathcal{K}}_{NM}^l = \mathcal{K}_{NM}^l \left( \mathcal{K}_{MM}^l \right)^{-1}$ and $
\widetilde{\mathcal{K}}_{NN}^l = \mathcal{K}_{NN}^l - \mathcal{K}_{NM}^l \left( \mathcal{K}_{MM}^l \right)^{-1} \mathcal{K}_{MN}^l$.

We will start by explicitly writing out Eq.~\eqref{eq:appxqfnL} and changing the order of integration:
\begin{equation} \label{eq:appxqfnLexplicit}
q(f_n^L) = \int q(f_M) \left[ \int \prod_{l=1}^{L}p(f_N^l|f_M^l;f_N^{l-1})  \prod_{n'\neq n} df_n^L \prod_{l=1}^{L-1} df_N^l \right] df_M.
\end{equation}
In the following we will only be concerned with the inner integral of the previous equation, which can also be written as
\begin{equation} \label{eq:appxqfnLinner}
 \int \left( \int p(f_N^L|f_M^L;f_N^{L-1})  \prod_{n'\neq n}  df^L_{n'} \right) \prod_{l=1}^{L-1} p(f_N^l|f_M^l;f_N^{l-1})  df_N^{l}.
\end{equation}
Here, the inner integral can be solved by exploiting the nice marginalisation property of multivariate Gaussians,
\begin{align}
\int p(f_N^L|f_M^L;f_N^{L-1}) \prod_{n'\neq n}  df^L_{n'} &= \int \gauss{ f_N^L }{ \widetilde{\mathcal{K}}_{NM}^L (f_N^{L-1}) f_M^L }{ \widetilde{\mathcal{K}}_{NN}^L (f_N^{L-1})} \prod_{n'\neq n}  df^L_{n'}\\
&= \gauss{ f_n^L }{ \widetilde{\mathcal{K}}_{nM}^L (f_N^{L-1}) f_M^L }{ \widetilde{\mathcal{K}}_{nn}^L (f_N^{L-1})},
\end{align}
where we explicitly marked the dependence of the $ \widetilde{\mathcal{K}}^L $ terms on the outputs $f_N^{L-1} $. While the $\widetilde{\mathcal{K}}^l_{nM}$ and $\widetilde{\mathcal{K}}^l_{nn}$ could in principle still depend on all the outputs of the previous layer, we see from their definitions after Eq.~\eqref{eq:appxreminder} that they in fact only depend on the marginals $ f_n^{L-1} $ and that therefore
\begin{equation}
\gauss{ f_n^L }{ \widetilde{\mathcal{K}}_{nM}^L (f_N^{L-1}) f_M^L }{ \widetilde{\mathcal{K}}_{nn}^L (f_N^{L-1})} = \gauss{ f_n^L }{ \widetilde{\mathcal{K}}_{nM}^L (f_n^{L-1}) f_M^L }{ \widetilde{\mathcal{K}}_{nn}^L (f_n^{L-1})} = p(f_n^L | f_M^L; f_n^{L-1}).
\end{equation}
Putting the last two equations together results in
\begin{equation}
\int p(f_N^L|f_M^L;f_N^{L-1}) \prod_{n'\neq n}  df^L_{n'} = p(f_n^L | f_M^L; f_n^{L-1}).
\end{equation}

We can continue with integrating out the $ f_N^l $ in Eq.~\eqref{eq:appxqfnLinner} in the same fashion (noting at every layer that we can not marginalise out $ f_n^l $ as those are inputs to kernels), arriving at
\begin{equation}
\int \prod_{l=1}^{L}p(f_N^l|f_M^l;f_N^{l-1})  \prod_{n'\neq n} df_n^L \prod_{l=1}^{L-1} df_N^l = \int \prod_{l=1}^L p(f_n^l | f_M^l; f_n^{l-1}) df_n^{1}\cdots df_n^{L-1}.
\end{equation}
Plugging this back into Eq.~\eqref{eq:appxqfnLexplicit} and changing the order of integration once again, yields
\begin{equation}\label{eq:appxqfnLproof2}
q(f_n^L) = \int \left[ \int q(f_M)\prod_{l=1}^L p(f_n^l | f_M^l; f_n^{l-1}) df_M \right] df_n^{1}\cdots df_n^{L-1},
\end{equation}
which is exactly Eq.~\eqref{eq:appxqfnLproof}, the result that we set out to show.

\subsection{Difference between mean-field and fully-coupled} \label{sec:appxdifMFFC}
From the previous equation it is also possible to see why a proof as in Appx.~\ref{sec:appxproof} was not necessary for the MF DGP.
This is due to the form of the variational posterior over the inducing outputs, given by
\begin{equation}\label{eq:appxpost}
q(f_M) = \begin{cases}
\gauss{f_M}{\mu_M}{S_M} & \text{for the FC DGP,} \\
\prod_{l=1}^L\prod_{t=1}^{T_l} \gauss{f_M^{l,t}}{\mu_M^{l,t}}{S_M^{l,t}} & \text{for the MF DGP.}
\end{cases}
\end{equation}
Using $ q(f_M) $ from the MF DGP, which can also be written as $ q(f_M) = \prod_{l=1}^L q(f_M^l) $, the inner integral in Eq.~\eqref{eq:appxqfnLproof2} can be rewritten as the product of $ l $ integrals,
\begin{equation}
\int q(f_M)\prod_{l=1}^L p(f_n^l | f_M^l; f_n^{l-1}) df_M =  \prod_{l=1}^L \int q(f_M^l) p(f_n^l | f_M^l; f_n^{l-1}) df_M^l,
\end{equation}
each being a standard integral in Gaussian calculus and the resulting formulas are given in Eqs.~\eqref{eq:dgpsamplingmf} and \eqref{eq:dgpsamplingfc}. In contrast, a fully coupled multivariate Gaussian can not be written as such a product, which is why the rather straightforward solution presented above is not possible in our case and the proof in Appx.~\ref{sec:appxproof} is needed.

%% file: subfiles/appendix_theory_part3.tex
\section{Linear algebra to speed up the code} \label{sec:appxLA}
In this section we provide some guidance through the linear algebra that is exploited in our code to speed up or vectorise calculations.
We will focus only on the most expensive terms, i.e., the off-diagonal covariance term $ \widetilde{S}_n^{l,1:l-1} $ and how to deal with $ \left(\widetilde{S}_n^{1:l-1,1:l-1}\right)^{-1} $, which are both needed to calculate $ \hat{\mu}^l_n $ and $ \hat{\Sigma}^l_n $ in Eqs.~\eqref{eq:appxmuhatn} and \eqref{eq:appxSigmahatn}, respectively. First, we show how to deal with the FC DGP and afterwards how the sparsity of $ S_M $ for the STAR DGP can be used.
The linear algebra that can be exploited for all the other terms, e.g.~the KL-divergence, will be provided along with the code.
Our implementation is in GPflow \cite{matthews2017gpflow} which provides all the functionalities that are necessary to deal with GPs in Tensorflow \cite{abadi2016tensorflow}.

Before we start we will repeat some formulas for convenience:
\begin{alignat}{2} \label{eq:appxmuhatn}
&\hat{\mu}_n^l &&= \widetilde{\mu}_n^l + \widetilde{S}_n^{l,1:l-1} \left( \widetilde{S}_n^{1:l-1,1:l-1} \right)^{-1} ( f_n^{1:l-1} - \widetilde{\mu}_n^{1:l-1}), \\ \label{eq:appxSigmahatn}
&\hat{\Sigma}_n^l &&= \widetilde{S}_n^{ll} - \widetilde{S}_n^{l,1:l-1} \left( \widetilde{S}_n^{1:l-1,1:l-1} \right)^{-1} \widetilde{S}_n^{1:l-1,l},
\\ \label{eq:appxStilden}
&\widetilde{S}_n^{ll'} &&=  \left(\widetilde{\mathcal{K}}^l_{Mn}\right)^\top S_M^{ll'} \widetilde{\mathcal{K}}^{l'}_{Mn}, \qquad \text{ if } l \neq l'.
\end{alignat}
Additionally, here is a more explicit definition of our notation of the covariance matrix $ S_M $:
\begin{equation}\label{eq:Sm}
S_M = 
\begin{pmatrix}
S_M^{11} & \cdots & S_M^{1L}\\
\vdots & \ddots & \vdots\\
S_M^{L1} & \cdots & S_M^{LL}
\end{pmatrix},\qquad
S_M^{ll'} =
\begin{pmatrix}
\left(S_M^{ll'}\right)_{11} & \cdots & \left(S_M^{ll'}\right)_{1T_{l'}}\\
\vdots & \ddots & \vdots\\
\left(S_M^{ll'}\right)_{T_l1} & \cdots & \left(S_M^{ll'}\right)_{T_lT_{l'}}
\end{pmatrix},
\end{equation}
where $ S_M $, $ S_M^{ll'} $, and $ \left(S_M^{ll'}\right)_{tt'} $ are matrices of size $M T \times M T $ (where $ T = \sum_{l=1}^{L} T_l $), $ MT_l\times MT_{l'} $, and $ M \times M $, which store the covariances of the inducing outputs between all inducing points, only those between layer $ l $ and $ l' $, and only those between the $ t $-th task in layer $ l $ and the $ t' $-th task in layer $ l' $, respectively.
In order to ensure that $ S_M $ is a valid covariance matrix (positive definite) we will numerically only work with its Cholesky decomposition $ L_S $ (s.t. $ S_M = L_SL_S^\top $), which is a lower triangular matrix.
Wherever possible we will want to avoid actually computing $ S_M $ and instead calculate all quantities from $ L_S $ directly.

\subsection{Fully coupled DGP}
\paragraph{Off-diagonal covariance terms} 
The terms \begin{equation}\label{eq:Snlsub}
\widetilde{S}_n^{l,1:l-1} = \begin{pmatrix}
\widetilde{S}_n^{l1} & \widetilde{S}_n^{l2} & \cdots & \widetilde{S}_n^{l,l-1}
\end{pmatrix},
\end{equation}
which are of size $ T_l \times \sum_{l'=1}^{l-1} T_{l'} $, have to be calculated for $ l=1,\dots,L $ and for $ n=1,\dots,N $.
As the number of layers $ L $ is in practice rather small, we will calculate all the individual matrices $ \widetilde{S}_n^{ll'} $ in a loop and concatenate them at the end, while we want to avoid a loop over $ N $.
Using Eq.~\eqref{eq:appxStilden} and that $\widetilde{\mathcal{K}}^{l}_{Mn} = \mathbb{I}_{T_l} \otimes \widetilde{K}^{l}_{Mn} $, we see that (for an example with $ T_l,T_{l'} = 2 $)
\begin{equation}\label{eq:Snll'}
\widetilde{S}_n^{ll'} =
\begin{pmatrix}
\left(\widetilde{K}^{l}_{Mn}\right)^\top (S_M^{ll'})_{11} \widetilde{K}^{l'}_{Mn} &
\left(\widetilde{K}^{l}_{Mn}\right)^\top (S_M^{ll'})_{12} \widetilde{K}^{l'}_{Mn}\\
\left(\widetilde{K}^{l}_{Mn}\right)^\top (S_M^{ll'})_{21} \widetilde{K}^{l'}_{Mn} &
\left(\widetilde{K}^{l}_{Mn}\right)^\top (S_M^{ll'})_{22} \widetilde{K}^{l'}_{Mn}
\end{pmatrix}.
\end{equation}
Writing $ \widetilde{S}_n^{ll'} $ in this way has two advantages:
Firstly, actually performing the multiplication $ \widetilde{\mathcal{K}}_{nM}^l S_M^{ll'} \widetilde{\mathcal{K}}_{Mn}^{l'} $ is extremely inefficient as the $\widetilde{\mathcal{K}}_{Mn}^{l}$ are block diagonal, which we resolved in this formulation.
Secondly, we note that exactly the same operation, i.e., multiplying from left and right by $\left(\widetilde{K}^{l}_{Mn}\right)^\top $ and $ \widetilde{K}^{l'}_{Mn} $, respectively, has to be performed on all $ T_l T_{l'} $  blocks of size  $ M\times M $.
This can be exploited since tensorflow has an inbuilt batch mode for most of its matrix operations.
In the following we will first show how the relevant block $ S_M^{ll'} $ can be efficiently obtained and afterwards show how to deal with the batch matrix multiplication for all $ n=1,\dots,N $.

Let us consider an example with three layers ($ L=3 $), the resulting covariance matrix and its Cholesky decomposition:
\begin{equation}\label{eq:traceSM}
S_M = 
\begin{pmatrix}
S_M^{11} & S_M^{12} & S_M^{13}\\
S_M^{21} & S_M^{22} & S_M^{23}\\
S_M^{31} & S_M^{32} & S_M^{33}
\end{pmatrix}
= L_S L_S^\top =
\begin{pmatrix}
L_S^{11} & 0 & 0\\
L_S^{21} & L_S^{22} & 0\\
L_S^{31} & L_S^{32} & L_S^{33}
\end{pmatrix}
\begin{pmatrix}
\left(L_S^{11}\right)^\top & \left(L_S^{21}\right)^\top & \left(L_S^{31}\right)^\top\\
0 & \left(L_S^{22}\right)^\top & \left(L_S^{32}\right)^\top\\
0 & 0 & \left(L_S^{33}\right)^\top
\end{pmatrix}.
\end{equation}
From this we can read off formulas for the blocks of $ S_M $, e.g., $ S_M^{32} = \begin{pmatrix} L_S^{31} & L_S^{32} \end{pmatrix}
\begin{pmatrix} L_S^{21} & L_S^{22} \end{pmatrix}^\top $, which in general can be written as
\begin{equation}\label{eq:SMll'}
S_M^{ll'} = L_S^{l,1:l'}\left(L_S^{l',1:l'}\right)^\top,
\end{equation}
where we exploited that we only need $ S_M^{ll'} $ for $ l'<l $ (the formula above is not valid for $ l' \geq l $). In this way we avoided calculating unnecessary matrix multiplications involving zero blocks.

Avoiding the loop over $ N $ requires a bit more linear algebra:
For this we note that e.g.~the element$ \left(\widetilde{S}_n^{ll} \right)_{11}$ can be seen as the $ n $-th diagonal element of the $ N \times N $ matrix
$\left(\widetilde{K}^{l}_{MN}\right)^\top (S_M^{ll'})_{11} \widetilde{K}^{l'}_{MN}$. Fully calculating this matrix is obviously very inefficient as we only need its diagonal elements. For this we use that, generally, for $ q \times p $ matrices $ A,\ C^\top $ and $ q \times q  $ matrices $ B $
\begin{equation}\label{eq:diagtrick}
\text{diag}\left(C^\top BA\right) = \text{column\_sum}(C^\top \odot BA),
\end{equation}
where $ \odot $ denotes the elementwise matrix product. The formula can easily be proved by explicitly writing the matrix products as sums and comparing terms on both sides.
Using this on all the blocks of $ \widetilde{S}_n^{ll'}$ in Eq.~\eqref{eq:Snll'} in a batched form and reordering the obtained terms afterwards requires some reshaping, which is explained in the code.

The most expensive calculations for this term are obtaining $ S_M^{ll'} $ [Eq.~\eqref{eq:SMll'}], which is $\mathcal{O}(M^3T_lT_{l'}\sum_{l''=1}^{l'} T_{l''})$ and the multiplication of e.g.\ $ (S_M^{ll'})_{11} \widetilde{K}^{l'}_{MN} $ which has to be done for all $ T_lT_{l'} $ blocks of $ S_M^{ll'} $ and is therefore $\mathcal{O}(NM^2T_lT_{l'})$. Both of these operations have to be performed for all $ l' = 1, \dots, l-1 $ in Eq.~\eqref{eq:Snlsub} and also for all layers $ l=1,\dots,L $. The total computational cost of this term is therefore $\mathcal{O}(M^3 \sum_{l=1}^L T_l \sum_{l'=1}^{l-1} T_{l'} \sum_{l''=1}^{l'} T_{l''} + N M^2 \sum_{l=1}^L T_l \sum_{l'=1}^{l-1} T_{l'})$.

\paragraph{Dealing with the inverse covariance terms}
First of all, we will never actually calculate $ \left(\widetilde{S}_n^{1:l-1,1:l-1}\right)^{-1} $. We will only use (and update) the lower triangular Cholesky decomposition $ L_{Sn}^{1:l-1,1:l-1} $ defined by
\begin{equation}\label{eq:cholSn}
L_{Sn}^{1:l-1,1:l-1} \left(L_{Sn}^{1:l-1,1:l-1}\right)^\top = \widetilde{S}_n^{1:l-1,1:l-1}.
\end{equation}
This can be done since the inverse term only ever appears in the product $ \widetilde{S}_n^{l,1:l-1} \left(\widetilde{S}_n^{1:l-1,1:l-1}\right)^{-1} $, whose transpose can be efficiently obtained via the solution of two triangular systems (which is implemented as cholesky\_solve in tensorflow). For this we obviously need the Cholesky decomposition first. We will point out how this can be efficiently calculated in the following, taking the calculations for the second layer as an example:

After having calculated $ \hat{\mu}^2_n $ and $ \hat{\Sigma}^2_n $ [Eqs.~\eqref{eq:appxmuhatn} and \eqref{eq:appxSigmahatn}, respectively]
we necessarily still have the quantities $\widetilde{L}_{Sn}^{11}$ (passed on from the first layer calculations), $\widetilde{S}_n^{12}$, and $\widetilde{S}_n^{22}$ in memory.
Note that we therefore can completely build the block matrix $  \widetilde{S}_n^{1:2,1:2}  $ as
\begin{equation}\label{eq:cholup2}
\widetilde{S}_n^{1:2,1:2} =
\begin{pmatrix}
\widetilde{L}_{Sn}^{11} \left(\widetilde{L}_{Sn}^{11}\right)^\top & \widetilde{S}_n^{12} \\
\left(\widetilde{S}_n^{12}\right)^\top & \widetilde{S}_n^{22},
\end{pmatrix}
\end{equation}
from which we could in principle calculate the Cholesky factor directly.
A more efficient way is to assume a general block matrix form for the Cholesky factor,
\begin{equation}\label{eq:cholup3}
\widetilde{L}_{Sn}^{1:2,1:2} =
\begin{pmatrix}
A & 0\\
B & C
\end{pmatrix},
\end{equation}
and then by using Eq.~\eqref{eq:cholSn} and comparing the terms to Eq.~\eqref{eq:cholup2} finding formulas for the unknown terms and solve them.
These formulas are given by
\begin{align}\label{eq:cholup41}
AA^\top &= \widetilde{L}_{Sn}^{11} \left(\widetilde{L}_{Sn}^{11}\right)^\top, \\\label{eq:cholup42}
AB^\top &= \widetilde{S}_n^{12},
\\\label{eq:cholup43}
BB^\top + CC^\top &= \widetilde{S}_n^{22}.
\end{align}
From Eq.~\eqref{eq:cholup41} we recognize $ A = \widetilde{L}_{Sn}^{11} $. Next, we can use this and Eq.~\eqref{eq:cholup42} to find $ B^\top $ as the solution of the (triangular) system $ \widetilde{L}_{Sn}^{11}B^\top = \widetilde{S}_n^{12} $.
The last step is then to obtain $ C $ from the Cholesky decomposition of the matrix $ \widetilde{S}_n^{22} - BB^\top $, where we used Eq.~\eqref{eq:cholup43}.
Note that while we still have to do a Cholesky decomposition, the matrix that has to be decomposed is (especially for large $ l $) considerably smaller than the full matrix $ \widetilde{S}_n^{1:2,1:2} $ and the computation therefore much faster.
For general $ l $ we simply have to substitute $ \widetilde{L}_{Sn}^{11}\rightarrow \widetilde{L}_{Sn}^{1:l-1,1:l-1} $, $ \widetilde{S}_n^{12} \rightarrow \widetilde{S}_n^{1:l-1,l} $, and $ \widetilde{S}_n^{22} \rightarrow \widetilde{S}_n^{ll} $, the rest stays the same. Plugging the obtained results for $ A $, $ B $, and $ C $ in Eq.~\eqref{eq:cholup3} yields the required "updated" Cholesky factor that needs to be passed on for the calculations in the next layer.

Solving the three equations \eqref{eq:cholup41} -  \eqref{eq:cholup43} for layer $ l $ requires $\mathcal{O}(N(T_l^2 \sum_{l'=1}^{l-1} T_{l'} + T_l (\sum_{l'=1}^{l-1} T_{l'})^2 + T_l^3))$ computation time. Doing so for all layers (note that we do not have to do this for the last layer) is therefore $\mathcal{O}(N\sum_{l=1}^{L-1} (T_l^2 \sum_{l'=1}^{l-1} T_{l'} + T_l (\sum_{l'=1}^{l-1} T_{l'})^2 + T_l^3))$

\paragraph{Computational costs}
Since these were the most expensive terms for the FC DGP, the overall computational cost for evaluating the ELBO is therefore
\begin{equation}\label{eq:compcos}
\mathcal{O}\left(N \sum_{l=1}^L \left[M^2 T_l \sum_{l'=1}^{l-1} T_{l'} + T_l^2 \sum_{l'=1}^{l-1} T_{l'} + T_l (\sum_{l'=1}^{l-1} T_{l'})^2 + T_l^3 + \frac{M^3}{N} T_l \sum_{l'=1}^{l-1} T_{l'} \sum_{l''=1}^{l'} T_{l''} \right]\right).
\end{equation}

In order to get a better grasp at this we will take an example architecture with $ L $ layers and the same number $ T_l = \tau $ of GPs per layer except for the last layer in which there is only one GP ($ T_L=1 $). In such a case the computational cost simplifies to
\begin{equation}\label{eq:compcostfc}
\mathcal{O}(NM^2\tau^2L^2 + N \tau^3 L^3 + M^3\tau^3L^3),
\end{equation}
where we only kept the highest order terms.

\subsection{Stripes-and-Arrow DGP}
In the following we will briefly sketch the computational savings that can be achieved for the two terms discussed in the previous section, when the restricted covariance of the STAR DGP is used.
Note that in this case the architecture necessarily needs to be the one decribed in the example given in the previous paragraph.
The (potential) non-zero $ M\times M $ blocks of the covariance matrix can be described by $ \left(S_M^{ll'}\right)_{tt'} $, where $ t=t' $ (this captures the terms also present in the MF DGP plus the diagonal stripes) or $ l = L $ or $ l'=L $ (this captures the two sides of the arrowhead).
It is easy to see that this form directly translates to the Cholesky decomposition which we will exploit in the next section:
The lower diagonal $ L_S $ has (potential) non-zero $ M\times M $ blocks $ \left(L_S^{ll'}\right)_{tt'} $,  only if $ l\geq l' $ (lower diagonal) and if $ t=t' $ (diagonal and stripes) or $ l = L $ (arrowhead).

\paragraph{Computing the covariance matrix from its Cholesky decomposition}
We will start by describing how the relevant terms of $ S_M $ can be obtained from our chosen sparse representation of $ L_S $, where we summarise the non-zero terms in three arrays, $ ^\text{diag}L_S $, $ ^\text{stripes}L_S $, and $ ^\text{arrow}L_S $.
The array $ ^\text{diag}L_S $ contains the $ \tau (L-1) +1  $ lower diagonal $ M\times M $ blocks on the diagonal of $ L_S $, where we access the $ t $-th block in the $ l $-th layer, i.e., $\left(L_S^{ll}\right)_{tt}$, by $ ^\text{diag}L_S^{l,t} $.
The array $ ^\text{stripes}L_S $ contains the $ \tau \sum_{k=1}^{L-2}k= \frac{\tau}{2}(L-2)(L-1)  $ blocks of size $ M\times M $, which form the diagonal stripes of $ L_S $, where we access  $\left(L_S^{ll'}\right)_{tt}$ (where $ L > l > l' $) by $ ^\text{stripes}L_S^{l,l',t} $.
The remaining $ \tau (L-1)  $  blocks of size $ M \times M $ of the arrowhead are contained in $ ^\text{arrow}L_S $, where we access  $\left(L_S^{Ll}\right)_{1t}$ as $ ^\text{arrow}L_S^{l,t} $. See also Fig.~\ref{fig:covariance} for a depiction of the covariance matrix.

The different terms of $ S_M $ can then be obtained as listed below:
\begin{enumerate}[i)]
	\item Diagonal terms $ \left(S_M^{ll}\right)_{tt} $ with $ l < L $:
	\begin{equation}
	\left(S_M^{ll}\right)_{tt} =\ ^\text{diag}L_S^{l,t} \left(^\text{diag}L_S^{l,t}\right)^\top + \sum_{l' = 1}^{l-1}\  ^\text{stripes}L_S^{l,l',t}  \left(^\text{stripes}L_S^{l,l',t} \right)^\top 
	\end{equation}
	
	\item Diagonal term $ \left(S_M^{LL}\right)_{11}$:
	\begin{equation}
	\left(S_M^{LL}\right)_{11} =\ ^\text{diag}L_S^{L,1} \left(^\text{diag}L_S^{L,1}\right)^\top + \sum_{l = 1}^{L-1} \sum_{t = 1}^\tau\  ^\text{arrow}L_S^{l,t}  \left(^\text{arrow}L_S^{l,t} \right)^\top 
	\end{equation}
	
	\item Stripe terms $ \left(S_M^{ll'}\right)_{tt} $ with $ L > l > l' $ ($ l < l' $ obtained via transposing):
	\begin{equation}
	\left(S_M^{ll'}\right)_{tt} =\ ^\text{stripes}L_S^{l,l',t} \left(^\text{diag}L_S^{l',t}\right)^\top + \sum_{l'' = 1}^{l'-1}\  ^\text{stripe}L_S^{l,l'',t}  \left(^\text{stripes}L_S^{l',l'',t} \right)^\top 
	\end{equation}
	
	\item Arrow terms $ \left(S_M^{Ll}\right)_{1t} $:
	\begin{equation}
	\left(S_M^{Ll}\right)_{1t} = \ ^\text{arrow}L_S^{l,t} \left(^\text{diag}L_S^{l,t}\right)^\top + \sum_{l' = 1}^{l-1}\ ^\text{arrow}L_S^{l',t} \left(^\text{stripes}L_S^{l,l',t}\right)^\top
	\end{equation}	
\end{enumerate}
The vectorisation of these equations can be seen in the code. The stripe terms are the most expensive to compute since an individual term has computational cost $ \mathcal{O}(M^3l') $ which has to be done for $ l = 2,\dots, L-1 $ and for $ l'=1,\dots,l-1 $ and for all $ t=1,\dots,\tau $. Therefore calculating all stripe terms is $ \mathcal{O}(M^3L^3\tau) $ (where we only kept the highest order term).

\paragraph{Off-diagonal covariance terms}
As before with the FC DGP, the off-diagonal covariance terms $ \widetilde{S}_n^{ll'} $ will also be the most expensive to compute. From the general formula in Eq.~\eqref{eq:appxStilden} it is easy to see that $ \left(\widetilde{S}_n^{ll'}\right)_{tt'} $ is only non-zero, if the corresponding  $ \left(S_M^{ll'}\right)_{tt'} $ are non-zero. We showed in the previous paragraph how those can be calculated, so the only step that remains to be done is the equivalent of Eq.~\eqref{eq:Snll'}, where again Eq.~\eqref{eq:diagtrick} can be used.

Doing this for an individual $ M\times M $ block can be done in $ \mathcal{O}(NM^2) $ time. Since we have to do this for all $ \mathcal{O}(\tau L^2) $ blocks, the total computational cost for the off-diagonal covariance terms is $ \mathcal{O}(NM^2\tau L^2) $.

\paragraph{Dealing with the inverse covariance terms}
For calculating (or updating) the Cholesky decomposition of $ \widetilde{S}_n^{1:l-1,1:l-1} $ we could in principle use similar ideas as we used above for calculating $ S_M $ (since both have the same sparsity pattern). But as we saw in the previous section, this term only incurs computational costs of $ \mathcal{O}(N\tau^3L^3) $ even for the FC DGP, which is for our settings always the least expensive term. We therefore simply reuse the algorithm described in the previous section to deal with this term and live with the resulting computational costs.

\paragraph{Computational costs}

The total computational costs for calculating the ELBO for the STAR DGP are therefore
\begin{equation}\label{eq:compcostsa}
\mathcal{O}(NM^2\tau L^2 + N \tau^3 L^3 + M^3\tau L^3).
\end{equation}

%% file: subfiles/appendix_pseudocode.tex
\section{Pseudocode} \label{sec:pseudocode}

We summarise the algorithm for calculating the ELBO~\eqref{eq:elbodgpfc} in Algs.~\ref{alg:ELBO}, \ref{alg:samplelayer}, and \ref{alg:getStilde} and ~\ref{alg:getStildeAFC}.
Alg.~\ref{alg:ELBO} shows how the ELBO can be calculated, where we average over multiple samples to reduce noise obtained by sampling through the layers. 
Alg.~\ref{alg:samplelayer} zooms into a single layer of the DGP and differentiates the mean-field approach and ours:
All coupled DGP approaches compute additional blocks of the covariance matrix $\widetilde{S}_n $ (marked in orange in Alg.~\ref{alg:getStilde} for the fully-coupled DGP and in Alg.~\ref{alg:getStildeAFC} for the stripes-and-arrow approximation). These blocks lead to a dependency of the output of the current layer $f_n^l$ on its predecessors $f_n^{1:l-1}$ (marked in orange).

	
	\begin{minipage}{\columnwidth}
		\begin{algorithm}[H]
			\caption{ELBO for coupled DGP}
			\label{alg:ELBO}
			\begin{algorithmic}
				\State given minibatch $ x_{N_b} $, $ y_{N_b} $ of size $N_b$, and
				\State number of Monte Carlo repetitions $ R $
				\State $ \mathcal{L} = 0 $
				\For{$ n=1\dots N_b $}
				\State $ \text{list} = [x_n] $ \Comment{accumulates $ \widetilde{\mathcal{K}}^l_{nM}, \widetilde{\mu}_n^l,\widetilde{S}_n^{1:l,1:l}, f_n^l $}
				\For{$ r=1\dots R $}		
				\For{$ l=1\dots L $} 
				\State $\hat{\mu}_n^l, \hat{\Sigma}_n^l, \text{list}  = \text{sample\_layer}(l,\text{list})$ \Comment{Alg.~\ref{alg:samplelayer}}	
				\EndFor
				\State $\mathcal{L} \mathrel{+}= 
				\frac{N}{N_b S}   \text{var\_log\_likelihood}(y_n,\hat{\mu}_n^L, \hat{\Sigma}_n^L)$
				\EndFor
				\EndFor
				\State $\mathcal{L} \mathrel{-}= \text{KL\_term}()$
				\State \Return $\mathcal{L}$
				\Comment{ELBO from Eq.~\eqref{eq:elbodgpfc}}
			\end{algorithmic}
		\end{algorithm}
	\end{minipage}
	
	\begin{minipage}{\columnwidth}
		\begin{algorithm}[H]
			\caption{$\text{sample\_layer}(l,\text{list})$: Return $\hat{\mu}_n^l, \hat{\Sigma}_n^l$ and sample $ f_n^l $, update relevant quantities for later layers.}
			\label{alg:samplelayer}
			\begin{algorithmic}
				\State \Comment{list contains $ \widetilde{\mathcal{K}}^{1:l-1}_{nM}, \widetilde{\mu}_n^{1:l-1},\widetilde{S}_n^{1:l-1,1:l-1}, f_n^{1:l-1} $}
				\State $ \widetilde{\mu}_n^{l} = \text{get\_mu\_tilde}() $ \Comment{Definition in Thm.~\ref{theoremproof} }
				\State $ \widetilde{S}_n^{l,1:l} $ = $\text{get\_S\_tilde}(l,\widetilde{\mathcal{K}}^{1:l}_{nM})$ \Comment{Alg.~\ref{alg:getStilde}}		
				\State $\hat{\mu}_n^l = \widetilde{\mu}_n^{l} + \textcolor{orange}{\text{correct\_mu}(\text{list}, \widetilde{S}_n^{l,1:l-1})} $ \Comment{Eq.~\eqref{eq:muhatn}}
				\State $\hat{\Sigma}_n^l = \widetilde{S}_n^{ll} - \textcolor{orange}{\text{correct\_Sigma}(\text{list}, \widetilde{S}_n^{l,1:l-1})} $ \Comment{Eq.~\eqref{eq:Sigmahatn}}
				\State $ f_n^l = \text{sample\_multivariate\_gauss}(\hat{\mu}_n^l,\hat{\Sigma}_n^l) $
				\State list = append(list, $\widetilde{\mathcal{K}}^l_{nM}$, $ \widetilde{\mu}_n^l $, $\widetilde{S}_n^{l,1:l-1}$, $\widetilde{S}_n^{ll}$, $ f_n^l $)
				\State \Return $\hat{\mu}_n^l, \hat{\Sigma}_n^l$, list
				\Comment{Return to Alg.~\ref{alg:ELBO}}		
			\end{algorithmic}
		\end{algorithm}
	\end{minipage}
	
	\begin{minipage}{\columnwidth}
		\begin{algorithm}[H]
			\caption{$\text{get\_S\_tilde}(l,\widetilde{\mathcal{K}}^{1:l}_{nM})$: Calculate $ \widetilde{S}_n^{l,1:l-1} $ and $ \widetilde{S}_n^{l,l} $ according to their definitions in Thm.~\ref{theoremproof} for the fully-coupled DGP. \newline	\newline
			$ (S_M^{ll'})_{tt'} $ denotes the $ M\times M $ block in $ S_M $ that contains the covariances of the inducing outputs of the $ t $-th GP in the $ l $-th layer and the $ t' $-th GP in the $ l' $-th layer. Analogously for $ (\widetilde{S}_n^{ll'})_{tt'} $.}
			\label{alg:getStilde}
			\begin{algorithmic}
				\For{$ l'=1\dots l $}
				\For{$ t,t'=1\dots T_l,1\dots T_{l'} $}
				\If{$ l=l' $ and $ t = t' $}
				\State $(\widetilde{S}_n^{l,l})_{tt}= K^l_{nn} + \widetilde{K}^l_{nM} \left( (S_M^{ll})_{tt} - K^l_{MM} \right) \widetilde{K}^{l}_{Mn}$
				\textcolor{orange}{
					\Else
					\State$(\widetilde{S}_n^{l,l'})_{tt'}=  \widetilde{K}^l_{nM} (S_M^{ll'})_{tt'} \widetilde{K}^{l'}_{Mn}$}
				\EndIf
				\EndFor
				\EndFor
				\State \Return $ \widetilde{S}_n^{l,1:l} $
				\Comment{Return to Alg.~\ref{alg:samplelayer}}		
			\end{algorithmic}
		\end{algorithm}
	\end{minipage}

	\begin{minipage}{\columnwidth}
\begin{algorithm}[H]
	\caption{$\text{get\_S\_tilde}(l,\widetilde{\mathcal{K}}^{1:l}_{nM})$: Calculate $ \widetilde{S}_n^{l,1:l-1} $ and $ \widetilde{S}_n^{l,l} $ for the stripes-and-arrow DGP.}
	\label{alg:getStildeAFC}
	\begin{algorithmic}
		\For{$ l'=1\dots l $}
		\For{$ t,t'=1\dots T_l,1\dots T_{l'} $}
		\If{{$ l=l' $ \textbf{ and } $ t = t' $}}
		\State $(\widetilde{S}_n^{l,l})_{tt}= K^l_{nn} + \widetilde{K}^l_{nM} \left( (S_M^{ll})_{tt} - K^l_{MM} \right) \widetilde{K}^{l}_{Mn}$
		\textcolor{orange}{\ElsIf{{$ l=L $} \textbf{ or } {$ t=t'$}}
		\State $(\widetilde{S}_n^{l,l'})_{tt'}=  \widetilde{K}^l_{nM} (S_M^{ll'})_{tt'} \widetilde{K}^{l'}_{Mn}$}
		\EndIf
		\EndFor
		\EndFor
		\State \Return $ \widetilde{S}_n^{l,1:l} $
		\Comment{Return to Alg.~\ref{alg:samplelayer}}		
	\end{algorithmic}
\end{algorithm}
	\end{minipage}

%% file: subfiles/appendix_exp.tex
\section{Additional experimental details} \label{sec:appxexdet}
In the following, we describe the experimental details necessary to reproduce the results:
\begin{itemize}
\item \textbf{Data Normalization} Normalization of inputs and outputs to zero mean and unit variance based on the training data.
\item \textbf{Inducing Inputs} $M=128$ inducing inputs initialized with k-means.
\item \textbf{DGP architecture}  $L=3$ hidden layers with $\tau =5$ latent GPs each and principal components of the training data as mean function.
\item \textbf{Likelihood} Gaussian likelihood with initial variance $\sigma^2_0 = 0.01$.
\item \textbf{Kernel} RBF kernel with automatic relevance determination (initial lengthscale $l_0=1$, 
 initial variance $\sigma^2_0=1$).  
\item \textbf{Optimizer} Adam Optimizer~\cite{kingma2014adam} (number of iterations $nIter=20,000$,
mini-batch size $mbs=512$, number of Monte Carlo for each data points $R=5$) with exponentially decaying learning rate (learning rate $lr=0.005$,  
 steps $ s=1000$, rate $r=0.98$).   
\item \textbf{Early Stopping}
In our initial experiments, we experienced overfitting for the variational methods.
To  prevent this from happening, we used 10\% of the training data as a validation set. 
We performed early-stopping if the performance on the validation set decreased in five successive strips~\cite{prechelt1998early}.
We did neither use a hold-out dataset for the GP methods, as they have a much smaller number of hyperparameters, nor for the Hamiltonian Monte Carlo approaches, as the correlation between adjacent samples complicates defining a good early-stopping criterion.
\item \textbf{Comparability}
Besides the early stopping criterion, we used the same model architectures, hyperparameter initialisations and optimisation settings across all methods and all datasets.
\item \textbf{Runtime Assessment}
The runtime of the different approximations was assessed for a single gradient update averaged over 2,000 updates on a 6 core i7-8700 CPU.
\item \label{it:nat_grads} \textbf{Natural gradients} For some experiments we also employed natural gradients, meaning we trained the model with a mixture of a natural gradient optimiser on all variational parameters and the Adam optimiser on all other parameters, similarly as in Refs.~\cite{salimbeni2018natgrads,hebbal2019bayesian}.
We additionally used exponentially decaying learning rates (learning rate Adam (natural gradients) $lr=0.001\ (0.005)$, steps $ s=1000$, rate $r=0.99$).
\end{itemize}

\renewcommand{\thefigure}{S\arabic{figure}}
\renewcommand{\thetable}{S\arabic{table}}

\begin{table*}
	\caption{
		\textbf{Extrapolation behaviour on  UCI benchmark datasets.}
		We report marginal test log-likelihoods (the larger, the better) for various methods and various number of layers $L$ in the extrapolation setting. 
		We marked all methods in bold that performed better or as good as SGP in the interpolation and extrapolation scenario, i.e., we simultaneously also looked at Tab.~\ref{tab:interpolation} in Sec.~\ref{sec:experiments}. We additionally underlined those that are significantly better (non-overlapping confidence intervals) in at least one of the scenarios.
		Standard errors are obtained by repeating the experiment 10 times.
	}
	\setlength{\tabcolsep}{4pt}
	\begin{footnotesize}
		\begin{center}
			\begin{tabular}{l|c|ccc|cc|cc}
				\toprule
				Dataset  & SGP & 
				\multicolumn{3}{c}{SGHMC DGP} &  
				\multicolumn{2}{c}{MF DGP} &
				\multicolumn{2}{c}{STAR DGP}  
				\\
				(N,D) & L1 & L1 & L2 & L3 &   L2 & L3 & L2 & L3
				\\
				\midrule
				boston (506,13) & -3.49(0.23) & -3.57(0.16) & -3.64(0.11) & -3.64(0.08) &\textbf{ -3.36(0.17)} &\textbf{ -3.41(0.19)} & \textbf{-3.38(0.18)} &\textbf{ -3.38(0.18)}\\
				energy (768, 8) & -2.90(0.63) & -3.22(0.69) & -3.53(0.89) & -3.26(0.75) & -3.02(0.64) & -3.45(0.86) & -3.08(0.78) & -2.95(0.74)\\
				concrete (1030, 8) & -3.91(0.11) & -3.90(0.06) & -4.37(0.19) & -4.71(0.33) & \textbf{-3.76(0.10)} & \textbf{-3.71(0.09)} & \textbf{-3.79(0.08)} &\textbf{ \underline{-3.68(0.08)}}\\
				wine red (1599,11) & -1.01(0.02) & -1.15(0.02) & -1.22(0.03) & -1.08(0.05) & -1.02(0.02) & -1.01(0.02) & \textbf{-1.01(0.02)} &\textbf{ -1.01(0.02)}\\
				kin8nm (8192, 8) & 0.66(0.04) & \textbf{\underline{0.72(0.03)} }& \textbf{\underline{1.06(0.03)}} & \textbf{\underline{0.94(0.11)} }& \textbf{\underline{0.96(0.03)}} & \textbf{\underline{0.98(0.04)} }&\textbf{ \underline{0.98(0.02)} }&\textbf{\underline{0.94(0.03)}}\\
				power (9568, 4) & -3.44(0.29) & -4.27(0.41) & -4.19(0.38) & -4.09(0.35) & -3.81(0.30) & -3.82(0.31) & -3.95(0.33) & -3.82(0.27)\\
				naval (11934,16) & 3.20(0.32) & 2.83(0.09) & 3.16(0.14) & 3.18(0.12) & 2.33(0.43) & 2.26(0.37) & 2.20(0.22) & 2.95(0.27)\\
				protein (45730, 9) & -3.20(0.04) &\textbf{ \underline{-3.20(0.03)} }& \textbf{\underline{-3.17(0.02)}} & \textbf{\underline{-3.10(0.02)}} & -3.31(0.04) & -3.23(0.06) & -3.23(0.04) & \textbf{\underline{-3.19(0.05)}}\\
				\bottomrule
				
			\end{tabular}
		\end{center}
		\label{tab:extrapolation}
	\end{footnotesize}
\end{table*}

\begin{table*}
	\caption{
{\new{
		\textbf{Extrapolation behaviour: direct comparison of DGP methods (part 2).} This table complements Tab.~\ref{tab:DGP-test}.
		In the fourth and in the last column, the same quantities as in Tab.~\ref{tab:DGP-test} are shown (see there for a description), where NG marks a method trained with natural gradients and FC stands for fully-coupled.		
		The other columns (2,3, and 5), marked with (dif), show means and standard errors (over 10 repetitions) of the \emph{difference} in marginal test log likelihood averaged over all test points in a single train test split.
		In each column, we mark numbers in bold if the second method outperforms the first, and in italics if it is the other way around.
		For the (dif) columns that is the case if the numbers significantly differ from zero.
		Note that the methods trained with natural gradients perform worse in the extrapolation task than those trained with Adam (as can be seen in the last column), a phenomenon that will have to be looked at more closely in the future.
	}
}}
	\setlength{\tabcolsep}{4pt}
\begin{footnotesize}
	\begin{center}
		\begin{tabular}{ l  | c c c c c}
			\toprule
			Dataset &    MF vs.~STAR (dif) & SGHMC vs.~STAR (dif) & MF NG vs.~FC NG & MF NG vs.~FC NG (dif) & MF NG vs.~MF\\
			\midrule
			boston & \textbf{0.036(0.029)}  & \textbf{0.27(0.15)} &  \textbf{0.58(0.04)} & \textbf{0.303(0.108)} & \textbf{0.63(0.04)}\\
			energy  & \textbf{0.500(0.202)} & \textbf{0.31(0.16)} & \textbf{ 0.70(0.05)}  &  \textbf{0.343(0.231)} & \textbf{0.77(0.05)} \\
			concrete  &  0.036(0.060)  & \textbf{1.03(0.33)} & \textbf{0.56(0.02)} & \textbf{0.145(0.070)} & \textbf{0.61(0.02)}\\
			wine red    &  \textbf{0.004(0.003)}  & \textbf{0.07(0.05)} & \textbf{ 0.57(0.03)}  & \textbf{0.018(0.003)} & \textbf{0.57(0.03)} \\
			kin8nm  & \textit{-0.040(0.024)} & -0.00(0.11) & \textbf{0.59(0.02)} &  \textbf{0.028(0.016)} & \textit{0.44(0.03)}\\
			power  & 0.005(0.096) & \textbf{0.27(0.11)} & \textbf{0.68(0.03)} & \textbf{0.355(0.157)} & 0.52(0.05)\\
			naval  & \textbf{0.693(0.527)} & -0.22(0.23) &  \textit{0.24(0.07)} & \textit{-0.178(0.112)} & \textbf{0.65(0.07)} \\
			protein  &  \textbf{0.033(0.014)} & \textit{-0.10(0.03)}  &   0.50(0.01)& \textit{-0.016(0.009)} & \textit{0.47(0.02)} \\
			\bottomrule
		\end{tabular}
	\end{center}
\label{tab:DGP-test-add}
\end{footnotesize}
\end{table*}

\begin{figure}
	\centering
		\includegraphics[width=\textwidth]{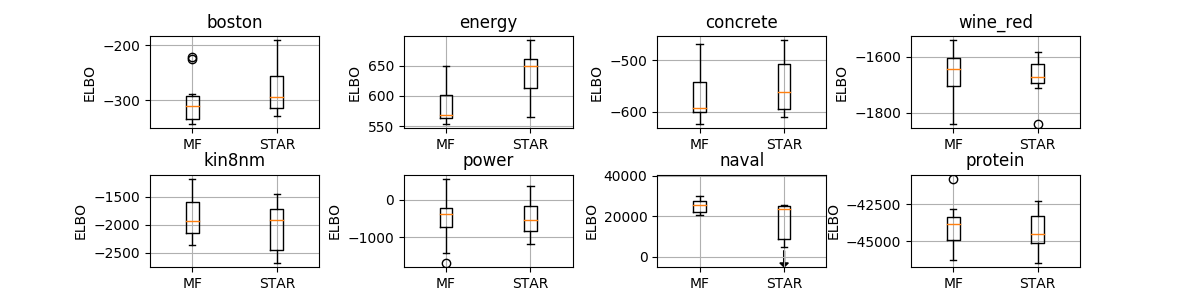}
\caption{\textbf{ELBO comparison.} We show boxplots of the ELBOs that correspond to the interpolation setting reported in Tab.~\ref{tab:interpolation} for all datasets and for 10 random splits, comparing the three layer versions of MF DGP and our STAR DGP. 
The performance increase of STAR DGP compared to MF DGP is the largest when the dataset is small (\textit{boston, energy, concrete}), while we observed a performance decrease on the dataset \textit{naval}.
For the latter, we also observed convergence difficulties in two runs, where the ELBOs are not plotted in the figure (indicated by the arrow).}
		\label{fig:elbo}
\end{figure}

\begin{table*}
	\caption{
		\textbf{ELBO comparison for fully-coupled DGP} We report  ELBOs (the larger, the better) for the mean-field (MF) and the fully-coupled (FC) method.
		We used our standard architecture with $ M=128 $, $\tau =5  $, and $ L=3 $ for both methods and trained them using natural gradients.
		Standard errors are obtained by repeating the experiment 10 times.
		We warm-started the optimisation of the fully-coupled method from the converged mean-field solution, using rather small learning rates (learning rate Adam (natural gradients) $lr=0.001\ (0.002)$, steps $ s=1000$, rate $r=0.95$, cf.~page~\pageref{it:nat_grads}) for a maximum of $ 7000 $ iterations.
		We marked the significant better performing (non overlapping standard errors) for each dataset in bold. Our structured approximation yields larger ELBOs for all datasets.}
	\setlength{\tabcolsep}{4pt}
	\begin{center}
		\begin{tabular}{l|cccccccc}
			\toprule
			Dataset & boston & energy & concrete & wine\_red & kin8nm & power &  naval & protein \\
			(N,D) & (506,13) & (768, 8) & (1030, 8) & (1599,11) & (8192, 8) & (9568, 4) & (11934,16) & (45730, 9) \\
			\midrule
			MF & -510(30) & 510(8) & -910(50) & -1648(8) & -2000(100) & -390(90) & 33000(600)  & -44040(140)\\
			FC & \textbf{-246(5)} & \textbf{600(20) }& \textbf{-500(10) }& \textbf{-1575(4)} & \textbf{-1290(70)} & \textbf{-10(50)} &\textbf{ 34600(500)} & \textbf{-42610(120)} \\
			\bottomrule
		\end{tabular}
	\end{center}	
	\label{tab:elbo-mf-fc}
\end{table*}

\begin{figure}
	\centering
		\includegraphics[width=0.48\textwidth]{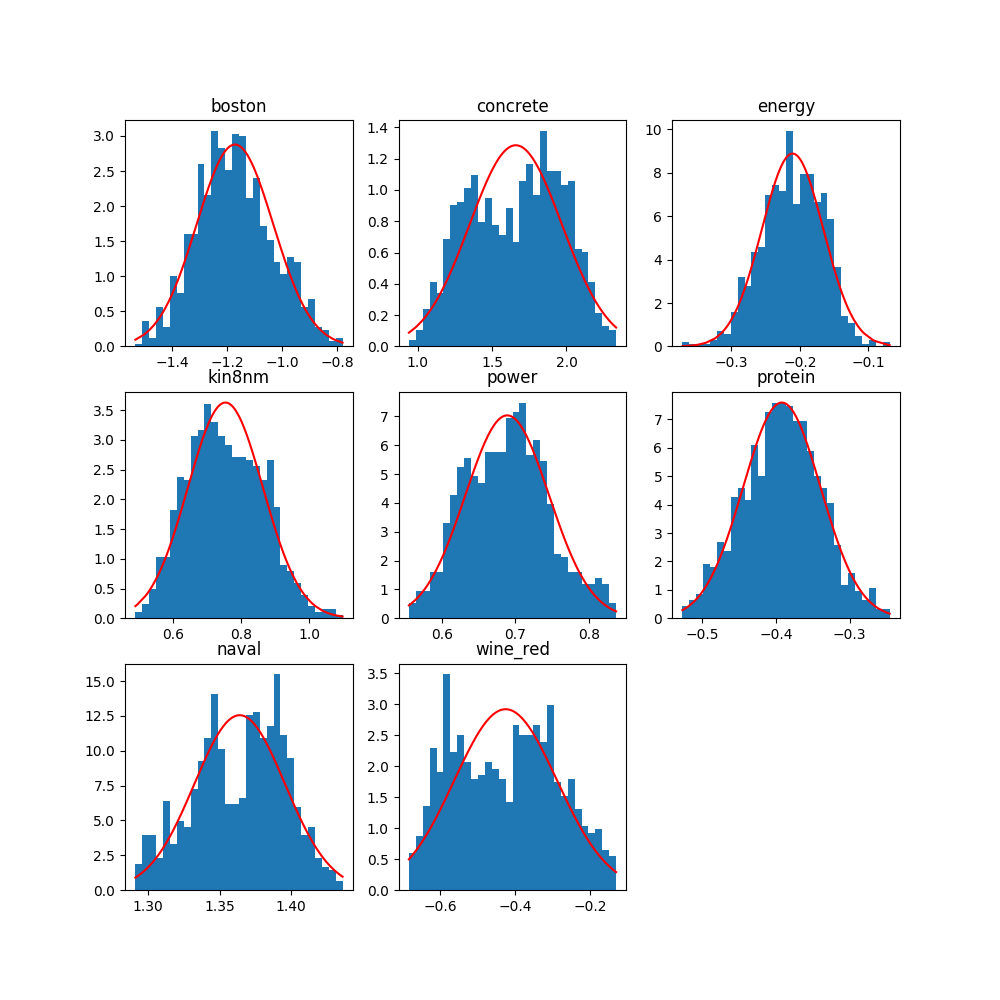}
			\includegraphics[width=0.48\textwidth]{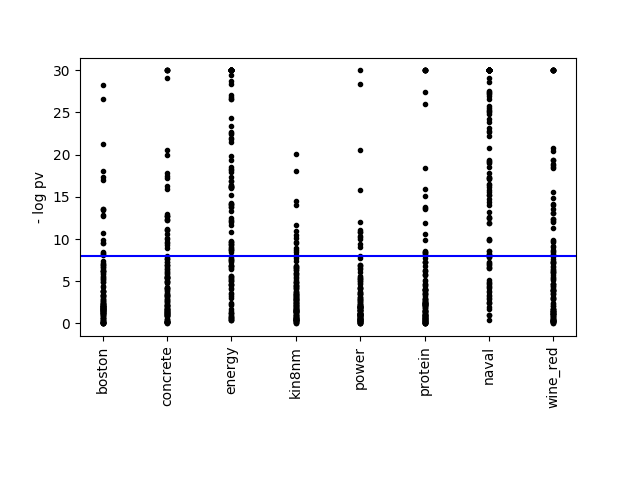}
\caption{
\textbf{Convergence of SGHMC.}
Left: 
Distribution over MC samples from a randomly chosen inducing output of a DGP with a single layer, equivalent to a SGP.
The red line indicates a Normal distribution fitted to the data.
Right: For each inducing output, we computed a p-value if its MC samples are normally distributed. The blue line shows the Bonferroni corrected significance threshold $\alpha=10^{-5}$.}
		\label{fig:sghmc}
\end{figure}

\begin{figure*}
		\centering
		\includegraphics[width=0.32\textwidth]{figures/runtime_M.png}
\includegraphics[width=0.32\textwidth]{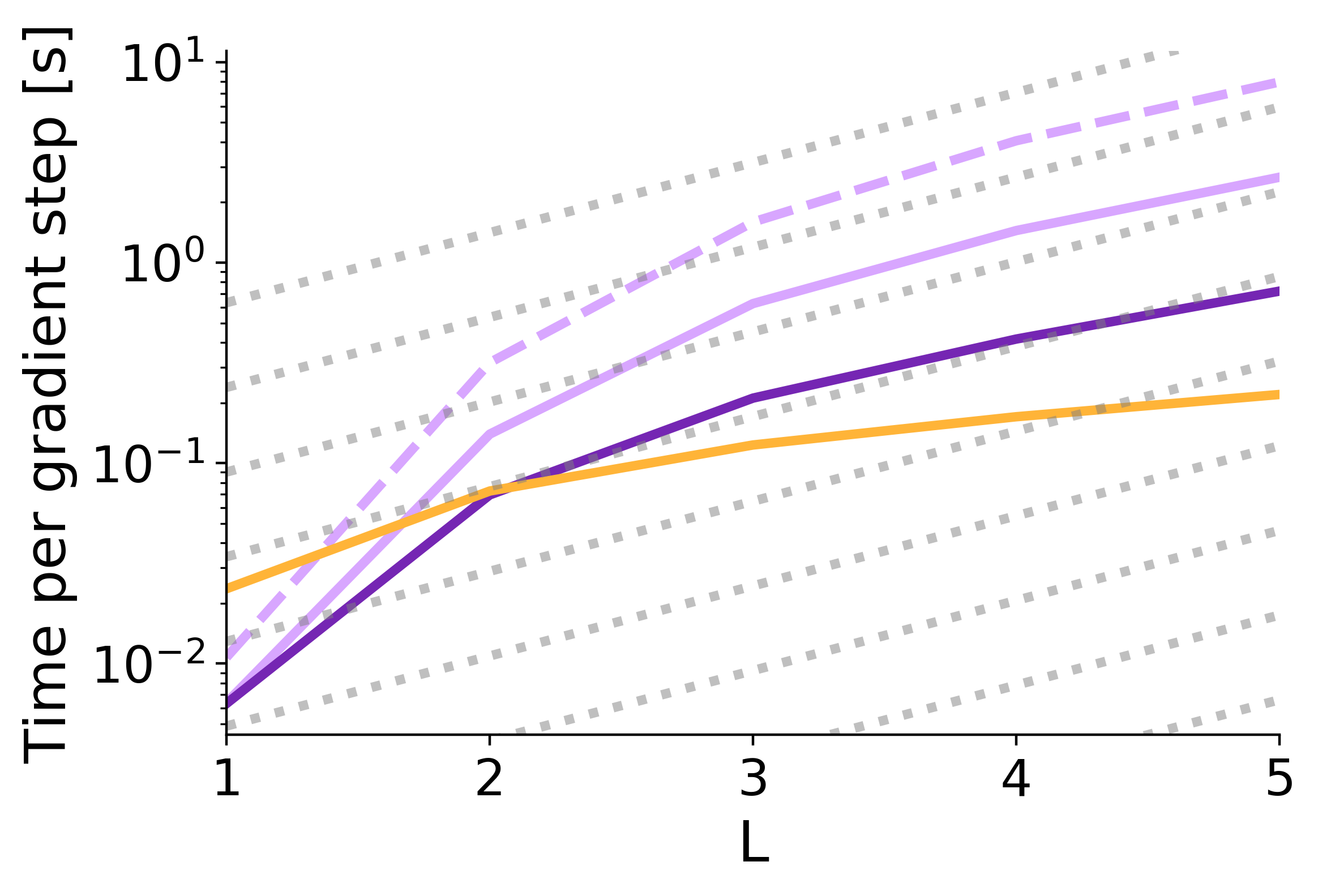}
\includegraphics[width=0.32\textwidth]{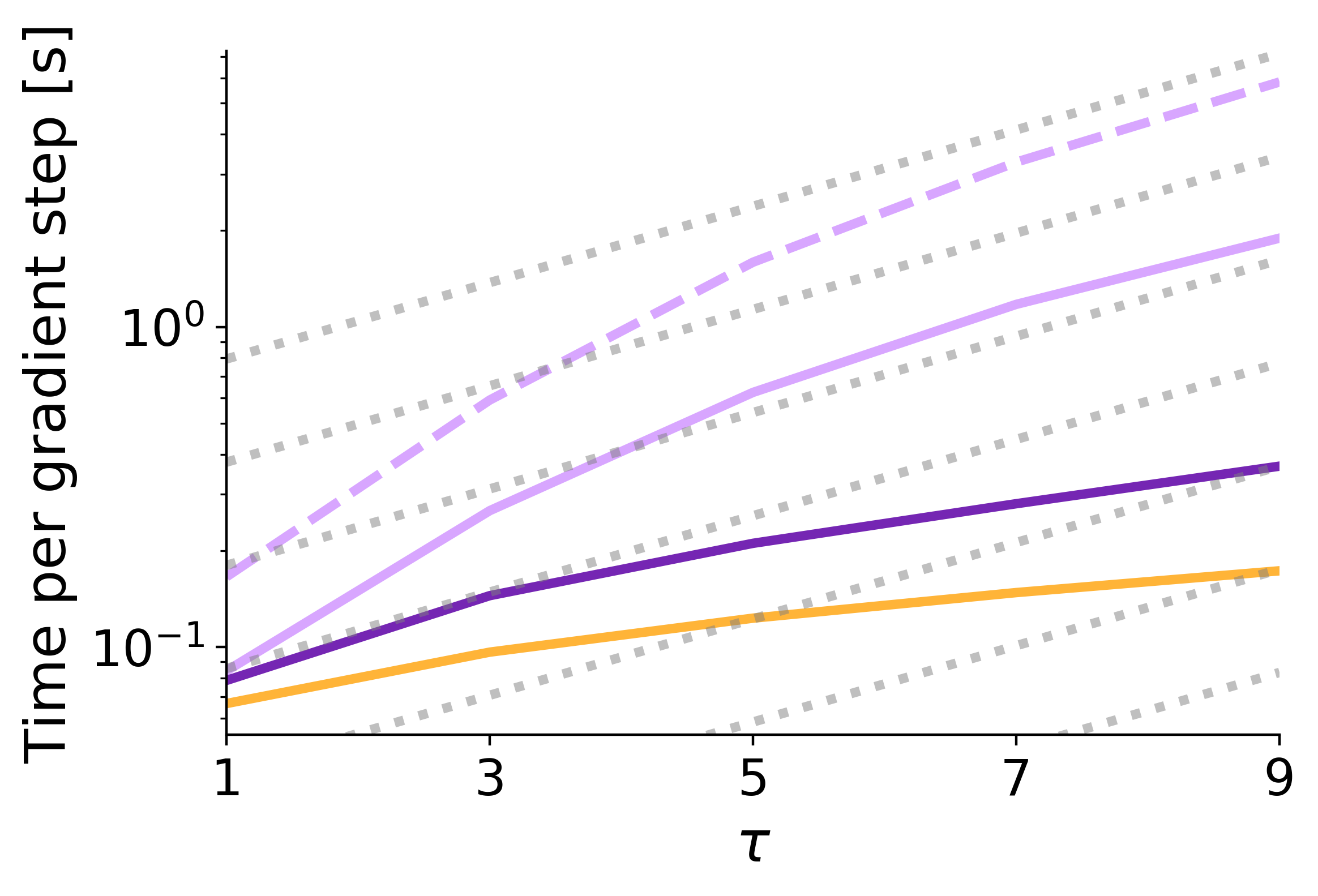}
\caption{
\textbf{Extended runtime comparison.}
We compare the runtime of our efficient stripes-and-arrow approximation (STAR DGP)  versus the fully coupled (FC DGP) and the 
mean-field approach (MF DGP) on the \textit{protein} UCI dataset. 
Shown is the runtime of one gradient step in seconds on a logarithmic scale as a function of the number of inducing points $M$, the number of layers $L$ and the number $\tau$ of latent GPs per intermediate layer (from left to right).
The dotted grey lines show the theoretical scaling of the runtime of the STAR DGP for the most important term $ \mathcal{O}(NM^2\tau L^2) $.
}
\label{fig:runtime2}
\end{figure*}

\begin{figure}
	\centering
	\includegraphics[width=0.29\textwidth]{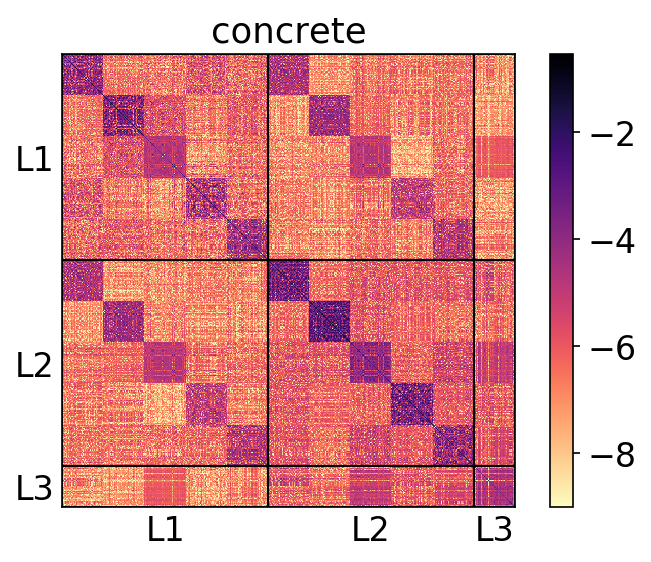}
	\includegraphics[width=0.29\textwidth]{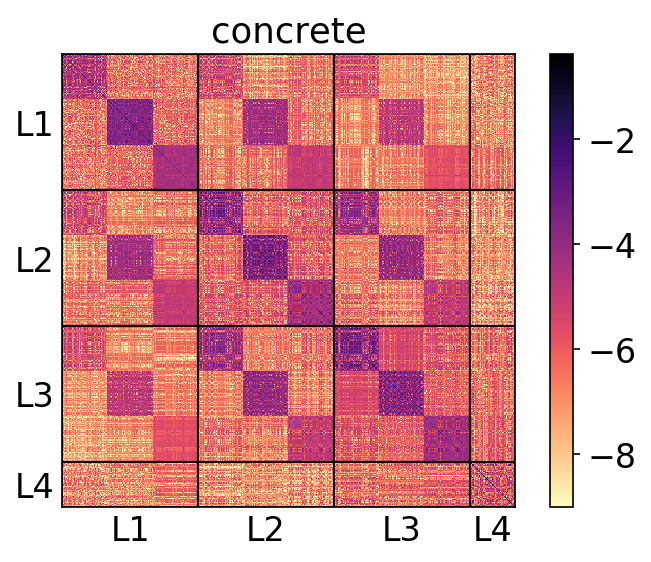}
	\includegraphics[width=0.29\textwidth]{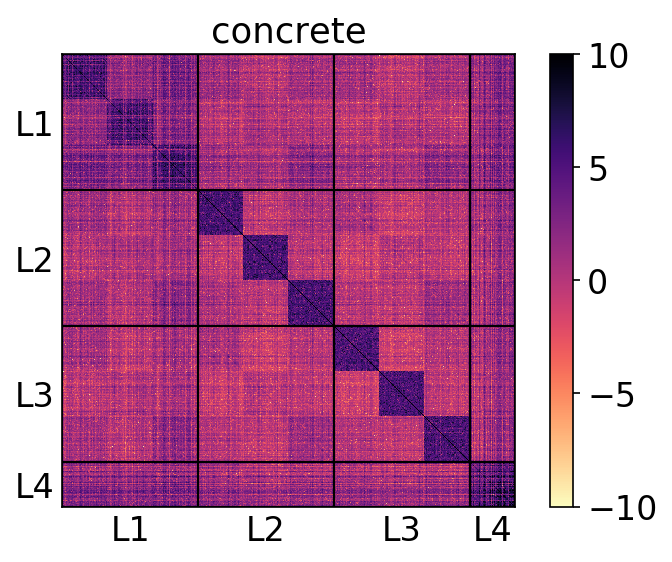}
	\includegraphics[width=0.29\textwidth]{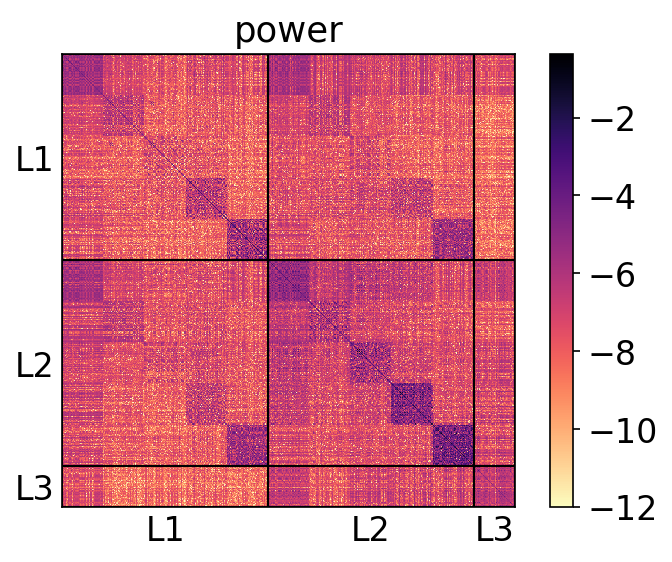}
	\includegraphics[width=0.29\textwidth]{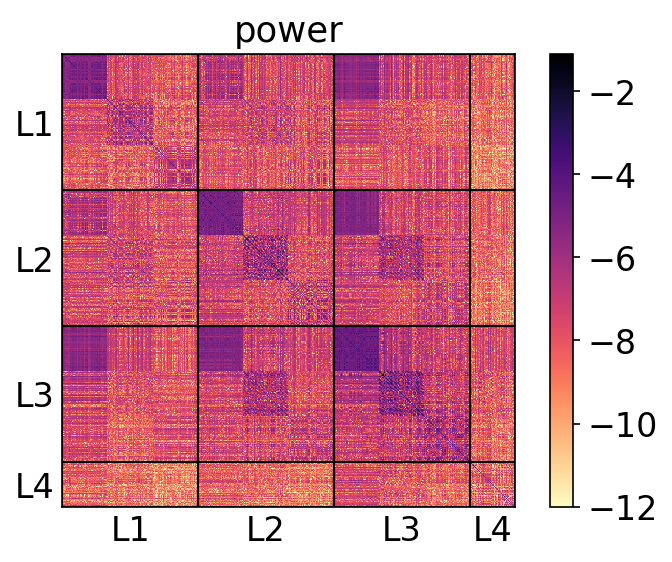}
	\includegraphics[width=0.29\textwidth]{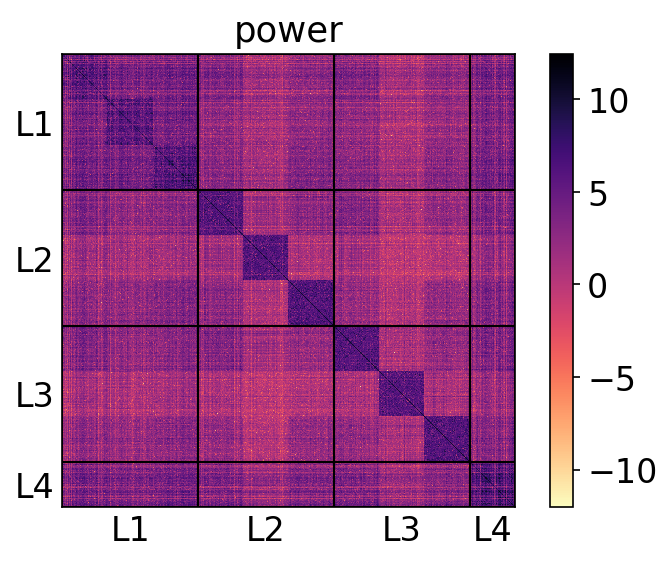}
	\includegraphics[width=0.29\textwidth]{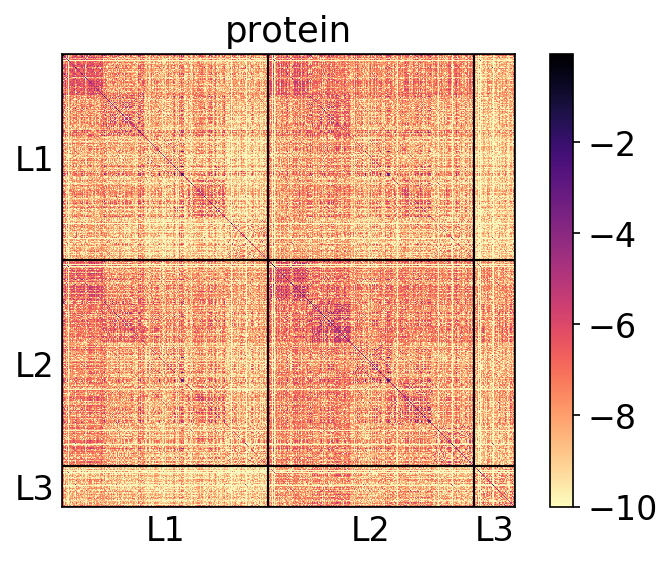}
	\includegraphics[width=0.29\textwidth]{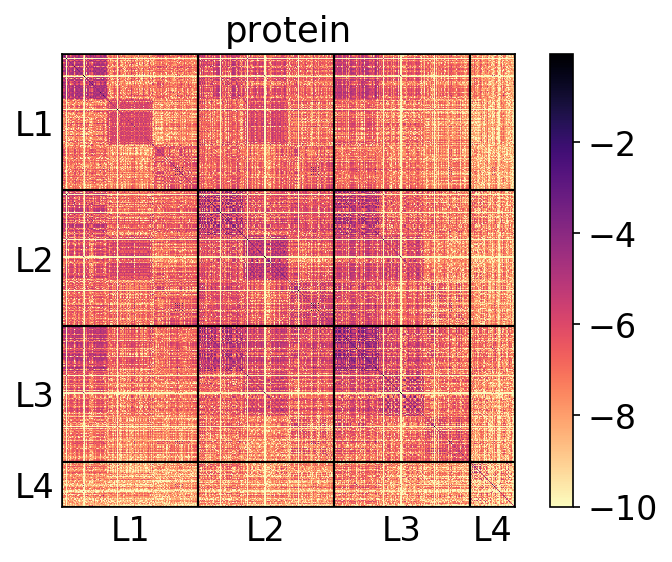}
	\includegraphics[width=0.29\textwidth]{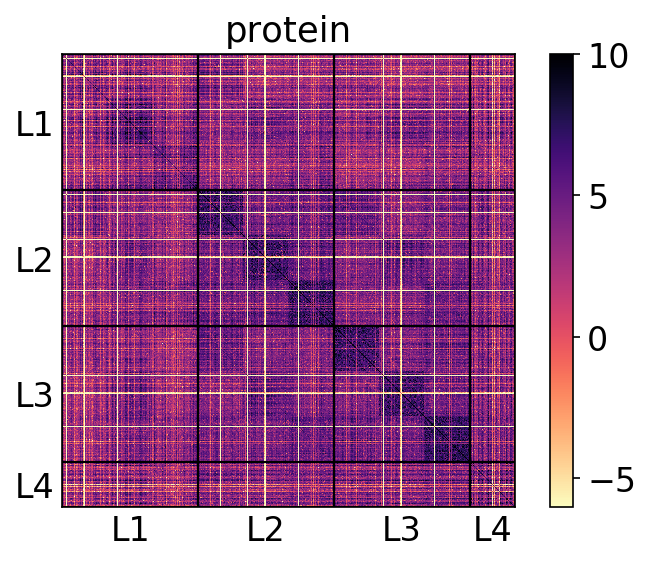}
	\caption{
		\textbf{Additional covariance and precision matrices.}
		Covariance/precision matrices after optimisation for three different UCI data sets. The first column depicts covariance matrices for our standard architecture, $ L=3, \tau = 5 $, while the second column depicts covariance matrices for $ L=4, \tau = 3 $. The third column depicts the precision matrices corresponding to the second column, i.e., the inverse matrices. 
Plotted are natural logarithms of the absolute values of the variational covariance/precision matrices over the inducing outputs.}
	\label{fig:add_cov}
\end{figure}